\documentclass[alpha-refs]{wiley-article}
\usepackage{amsmath}
\usepackage[utf8]{inputenc}
\usepackage{mathtools}
\usepackage{multirow}
\newcolumntype{P}[1]{>{\RaggedRight\arraybackslash}p{#1}}
\usepackage{siunitx}
\usepackage{ulem}
\usepackage{fancybox}
\usepackage{placeins}
\usepackage{wrapfig}
\usepackage{longtable}
\usepackage{graphicx}
\usepackage [utf8]{inputenc}
\usepackage{xcolor} 
\usepackage{comment}
\usepackage{natbib}
\usepackage{colortbl} 
\usepackage{pifont}   
\usepackage{array} 
\usepackage{hyperref}
\usepackage{xcolor}
\usepackage{tikz}
\usepackage{epstopdf}
\usepackage{fancyhdr}
\DeclareGraphicsExtensions{.tif}
\fancypagestyle{firstpage}{
  \fancyhf{} 
  \fancyhead[R]{Under Review} 
}

\AtBeginDocument{
  \setlength\abovedisplayskip{0pt}
  \setlength\belowdisplayskip{0pt}}
\usepackage[margin=0pt, font=small]{caption}
\usepackage[moderate]{savetrees}
\newcommand*\emptycirc[1][1ex]{\tikz\draw (0,0) circle (#1);} 
\newcommand*\halfcirc[1][1ex]{%
  \begin{tikzpicture}
  \draw[fill] (0,0)-- (90:#1) arc (90:270:#1) -- cycle ;
  \draw (0,0) circle (#1);
  \end{tikzpicture}}
\newcommand*\fullcirc[1][1ex]{\tikz\fill (0,0) circle (#1);} 
\newcolumntype{C}[1]{>{\centering\arraybackslash}p{#1}}
\newcommand{\blackcircled}[1]{\tikz[baseline=(char.base)]{
            \node[shape=circle,fill=black,text=white,draw,inner sep=1pt] (char) {#1};}}

\newcommand{\cmark}{\ding{51}}  
\newcommand{\xmark}{\ding{55}}

\begin{document}
\thispagestyle{firstpage}

\papertype{Advanced Review}
\paperfield{Journal Section}

\title{A Survey on Efficient Vision-Language Models}

\author[$\spadesuit$]{Gaurav Shinde$^\ddagger$}
\author[$\spadesuit$]{Anuradha Ravi}
\author[$\spadesuit$]{Emon Dey}
\author[$\spadesuit$]{Shadman Sakib}
\author[$\spadesuit$]{Milind Rampure}
\author[$\spadesuit$]{Nirmalya Roy}

\affil[$\spadesuit$]{Mobile Pervasive \& Sensor Computing Lab and Department of Information Systems, University of Maryland Baltimore County~(UMBC), Baltimore, Maryland, 21250, USA} 

\corraddress{{$^\ddagger$}Gaurav Shinde, \newline Mobile Pervasive \& Sensor Computing Lab and Department of Information Systems
University of Maryland Baltimore County (UMBC), Baltimore, Maryland, 21250, USA}
\corremail{gshinde1@umbc.edu}

\presentadd{Mobile Pervasive \& Sensor Computing Lab and Department of Information Systems, University of Maryland Baltimore County (UMBC), Baltimore, Maryland, 21250, USA}

\fundinginfo{This work has been partially supported by NSF CAREER Award \# 1750936, NSF REU Site Grant \# 2050999, NSF CNS EAGER Grant \# 2233879, and ONR Grant \# N00014-23-1-2119}
\runningauthor{Gaurav Shinde et al.}


\maketitle

\begin{abstract}
Vision-language models (VLMs) integrate visual and textual information, enabling a wide range of applications such as image captioning and visual question answering, making them crucial for modern AI systems. However, their high computational demands pose challenges for real-time applications. This has led to a growing focus on developing efficient vision-language models. In this survey, we review key techniques for optimizing VLMs on edge and resource-constrained devices. We also explore compact VLM architectures, frameworks and provide detailed insights into the performance-memory trade-offs of efficient VLMs. Furthermore, we establish a GitHub repository at \textcolor{blue}{\url{https://github.com/MPSC-UMBC/Efficient-Vision-Language-Models-A-Survey}} to compile all surveyed papers, which we will actively update. Our objective is to foster deeper research in this area.
\end{abstract}


\section{Introduction}
\label{sec:introduction}
Vision-Language Models (VLMs) have emerged as a response to the growing need for systems that can process and integrate visual and textual data effectively. The increasing availability of multimodal data across domains such as healthcare (medical images paired with diagnostic reports), autonomous systems (sensor feeds integrated with navigation commands), and social media (images combined with captions) highlighted the limitations of unimodal models, which struggled to connect visual content with its linguistic context. VLMs address this challenge by aligning vision and text in a shared representation space which enables advanced capabilities in tasks such as image captioning, cross-modal retrieval, visual question answering (VQA) and visual commonsense reasoning (VCR). Advances in deep learning architectures and the availability of large-scale multimodal datasets have further fueled the development of VLMs.

VLMs leverage diverse objectives to align and integrate multimodal data effectively, with approaches such as contrastive learning, masked modeling, and generative modeling playing a pivotal role. In contrastive-based VLMs, the goal is to assign low energy to consistent data pairs while penalizing unlikely combinations with higher energy. The learned energy function \( E_{\phi}(x) \) maps these data samples into a probability distribution using the Boltzmann formula:
\begin{equation}
P_{\phi}(x) = \frac{e^{-E_{\phi}(x)}}{\sum_x e^{-E_{\phi}(x)}}
\end{equation}
This ensures that samples with lower energy correspond to higher probabilities. The objective is to align \( P_{\phi}(x) \) the model's distribution with \( P_{T}(x) \) the target distribution. This optimization often involves the maximum likelihood estimation, with gradients computed over positive and negative samples, where negative samples are synthesized through methods like Markov Chain Monte Carlo (MCMC). Models like CLIP \cite{Radford2021} and SigLIP \cite{Zhai2023} demonstrate the effectiveness of contrastive learning by aligning visual and textual embeddings in a shared representation space, enabling robust performance across a wide range of multimodal tasks.
Masked modeling offers a different approach by masking parts of the input and training the model to predict the masked portion. Masked Language Modeling (MLM), for example, benefits from the transformer architecture by strategically dropping input tokens and predicting them, while Masked Image Modeling (MIM) employs similar principles for visual data. Frameworks like FLAVA \cite{Singh2022} and BEiT \cite{Bao2021} effectively utilize masked modeling to pre-train on vast multimodal datasets.
Generative models, on the other hand, extend the capabilities of VLMs by simultaneously learning contrastive and generative losses. Generative models are widely used for image captioning tasks. CM3Leon \cite{Yu2023CM3Leon} utilizes distinct image and text tokenizers to convert respective modalities into token sequences, which are subsequently processed by a transformer decoder. In contrast, Chameleon \cite{Team2024Chameleon} extends this framework by employing the same transformer model for processing both image and text tokens to ensure a unified and efficient design. Although widely applied to image captioning tasks, generative models can also be utilized for various downstream applications. For example, they can be employed for image classification tasks using Bayes' rule. The probability for a given label \( y \) given an image \( x \) can be expressed as:
\begin{equation}
P(y|x) = \frac{P(x|y)P(y)}{P(x)}
\end{equation}
To optimize computational resources and reduce training costs, VLMs often integrate pre-trained backbones such as Frozen \cite{Tsimpoukelli2021Frozen}, MiniGPT \cite{Zhu2023MiniGPT4EV}, or the Qwen series \cite{qwen2024qwen25technicalreport}. These pretrained components accelerate convergence and generalize effectively across tasks. VLM pre-training frameworks differ in architecture, ranging from two-tower VLMs with distinct encoders for image and text to one-tower VLMs that utilize unified networks to generate joint embeddings. These approaches enhance the efficiency of VLMs, making them useful for edge device applications.

\begin{figure}
    \centering
    \includegraphics[width=\textwidth]{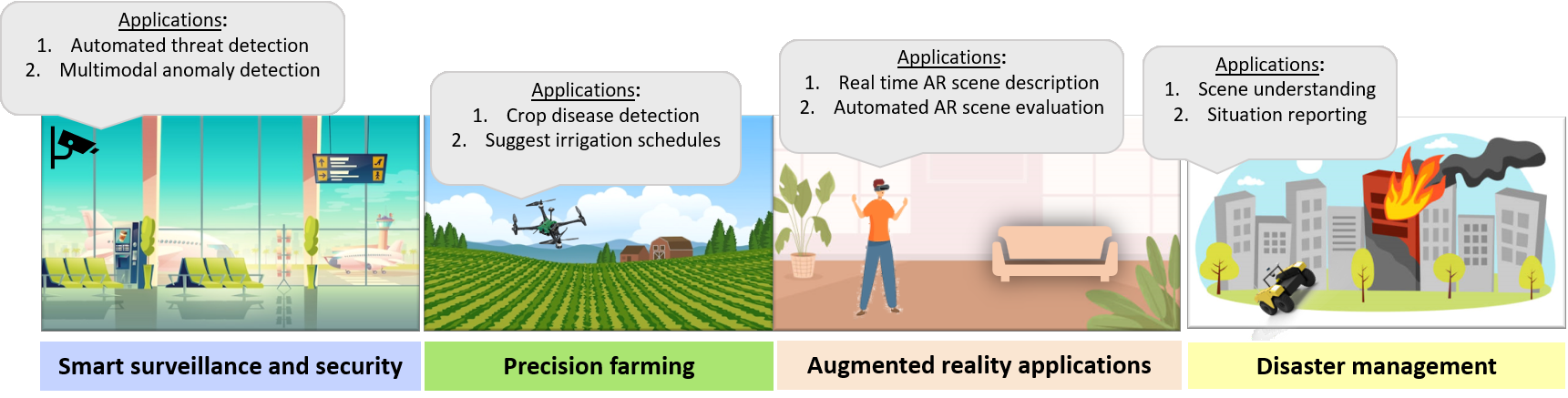}
    \caption{Key VLM applications highlighting the necessity of edge deployment.}
    \label{fig:VLMedgeApplications}
\end{figure}

The deployment of VLMs on resource-constrained devices addresses the growing need for real-time processing and privacy preservation by enabling on-device computations. Furthermore, edge deployment ensures seamless performance in environments with limited or unreliable network connectivity, making VLMs highly adaptable for applications like autonomous navigation and smart IoT systems (figure \ref{fig:VLMedgeApplications}).
As state-of-the-art VLMs grow in complexity to achieve higher performance, their memory footprint and inference latency have increased significantly. For example, the CLIP-B/16 model \cite{Liu2024} consists of 86.2M parameters for the image encoder and 63.4M parameters for the text encoder, making them unsuitable for deployment on edge devices such as the Jetson Nano (4 GB RAM, no dedicated GPU) or even the Jetson Xavier (8 GB RAM, 1 GPU). On a Jetson Nano, the limited RAM would cause frequent memory swaps, severely impacting latency and throughput, while on the Jetson Xavier, the GPU may still struggle to handle the large model's computational demands in real-time applications. These limitations emphasize the critical need for efficient VLMs that optimize memory usage and latency while maintaining competitive performance.

\textit{The main contributions of this survey are threefold}. \blackcircled{1} We present various techniques, including pre-deployment, efficient fine-tuning, and runtime optimizations, to enhance the efficiency of VLMs for resource-constrained devices. \blackcircled{2} We list state-of-the-art lightweight VLMs and discuss various frameworks. \blackcircled{3} We provide detailed insights into the performance-memory trade-off for certain VLMs using the techniques discussed.

\begin{figure}
    \centering
    \includegraphics[width=\textwidth]{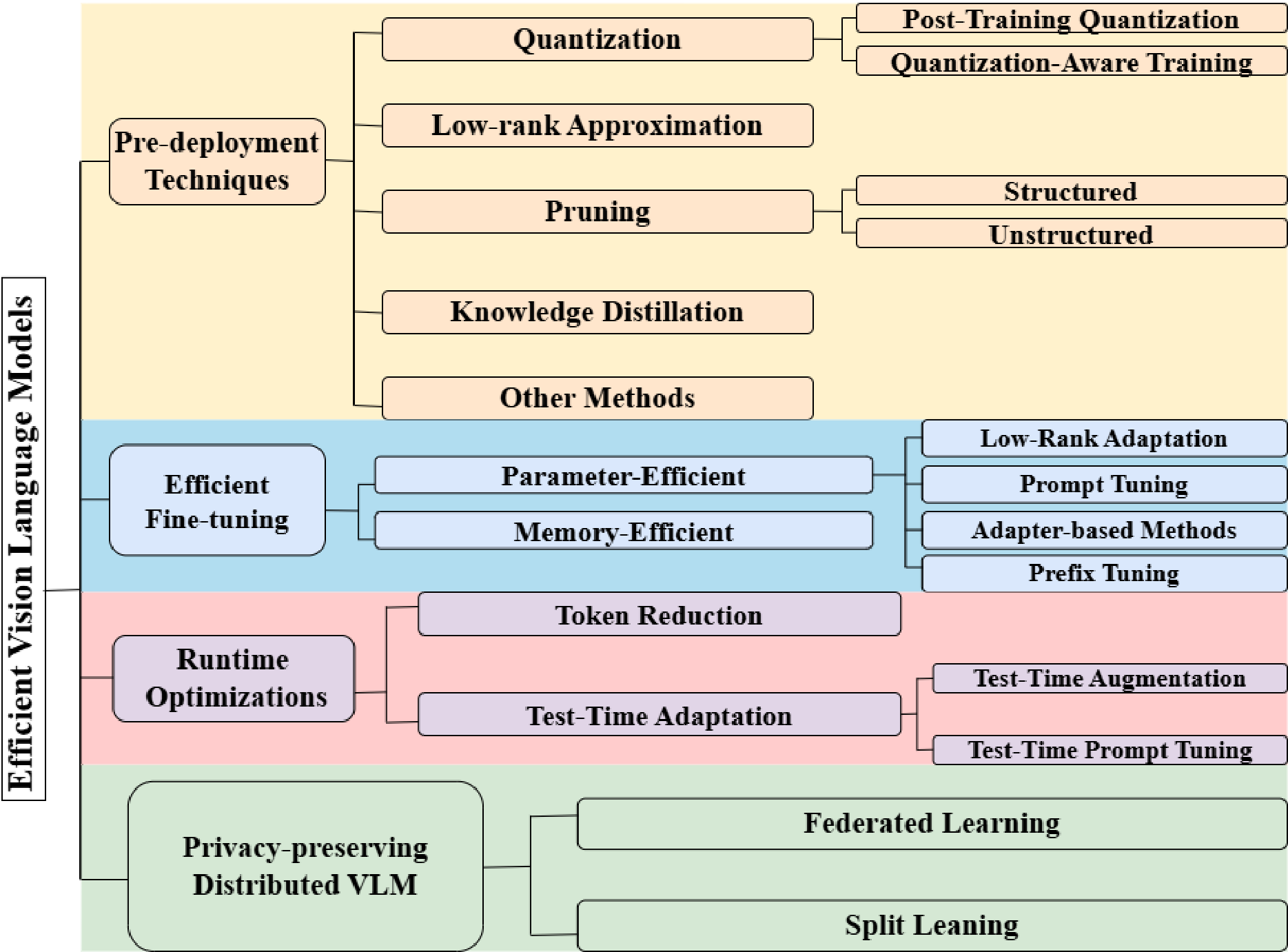}
    \caption{Taxonomy of efficient vision-language models (VLMs).}
    \label{fig:Taxonomy}
\end{figure}

This survey is structured around the taxonomy illustrated in Figure \ref{fig:Taxonomy}. While existing surveys such as \cite{ghosh2024exploringfrontiervisionlanguagemodels} primarily explore various VLM architectures and \cite{Du2022VL-PTM} delve into Vision-Language Pre-Trained Models (VL-PTMs) and \cite{Zhang2024KD} mention techniques like knowledge distillation and transfer learning, our work provides an in-depth analysis of efficient VLMs specifically tailored for edge and resource-constrained devices. Table \ref{tab:survey_comparison_table} compares our survey with similar ones. To ensure a thorough review we sourced relevant papers from leading conferences and workshops using platforms such as Google Scholar, DBLP and ResearchGate. Keywords such as "VLM quantization," "VLM pruning," "VLM finetuning techniques," "VLM knowledge distillation," and "VLM runtime optimizations" guided our literature search, enabling a targeted exploration of this rapidly evolving field.

The rest of the survey is organized as follows. Section \ref{predeployment} covers pre-deployment techniques for efficient VLMs, while Sections \ref{efficientfinetuning} and \ref{runtime optimizations} detail fine-tuning and runtime optimization methods. Section \ref{privacyprservingdistributedvlm} explores distributed VLMs. Section \ref{Efficient VLMs} lists state-of-the-art, efficient VLMs and also details various VLM frameworks and libraries. Section \ref{insights} provides insights into the accuracy vs. efficiency trade-offs for VLMs. Section \ref{applications} outlines application avenues, Section \ref{challenges} discusses challenges and future directions, and Section \ref{conclusionfuturedirections} concludes the paper.

We also create a \textit{\textbf{GitHub repository}} to compile the papers featured in this survey: \textcolor{blue}{\url{https://github.com/MPSC-UMBC/Efficient-Vision-Language-Models-A-Survey}}. It will be actively maintained with updates on emerging research.

\begin{table}[!htbp]
\centering
\caption{Comparison of our survey with similar ones; \fullcirc means full coverage of the topic, \halfcirc means partial coverage of the topic, and \emptycirc means no coverage.
\label{tab:survey_comparison_table}}
\scalebox{1}{
    \scriptsize
    \setlength{\tabcolsep}{2pt}  
    \begin{tabular}{|C{2cm}|C{2cm}|C{2cm}|C{1.7cm}|C{1.5cm}|C{1.2cm}|C{2cm}|}
    \hline
    \textbf{Reference} & \textbf{Pre-deployment Techniques} & \textbf{Efficient Fine-tuning Techniques} & \textbf{Runtime Optimization} & \textbf{Security and Privacy} & \textbf{Applications} &\textbf{Memory vs. Performance Insights} \\
    \hline
    \cite{Survey1} & \emptycirc & \fullcirc & \emptycirc & \fullcirc & \emptycirc & \emptycirc\\ 
    \hline
    \cite{Survey2} & \halfcirc & \fullcirc & \emptycirc & \emptycirc & \fullcirc & \emptycirc \\
    \hline
    \cite{survey3} & \halfcirc & \fullcirc & \halfcirc & \fullcirc & \fullcirc & \emptycirc \\
    \hline
    \cite{Survey4} & \halfcirc & \halfcirc & \emptycirc & \emptycirc & \halfcirc & \emptycirc \\
    \hline
    \textbf{\underline{Ours}} & \fullcirc & \fullcirc & \fullcirc & \fullcirc & \fullcirc & \fullcirc\\
    \hline
    \end{tabular}
}
\end{table}

\section{Pre-deployment Techniques}
\label{predeployment}
Pre-deployment techniques refer to optimization methods applied to models before deployment to improve their efficiency while maintaining accuracy. In our survey, we focus on techniques such as quantization, low-rank approximation, pruning, and knowledge distillation. These methods are particularly suited for edge devices as they reduce computational and memory requirements. 

\subsection{Quantization}
\begin{figure}
    \centering
    \includegraphics[width=\textwidth]{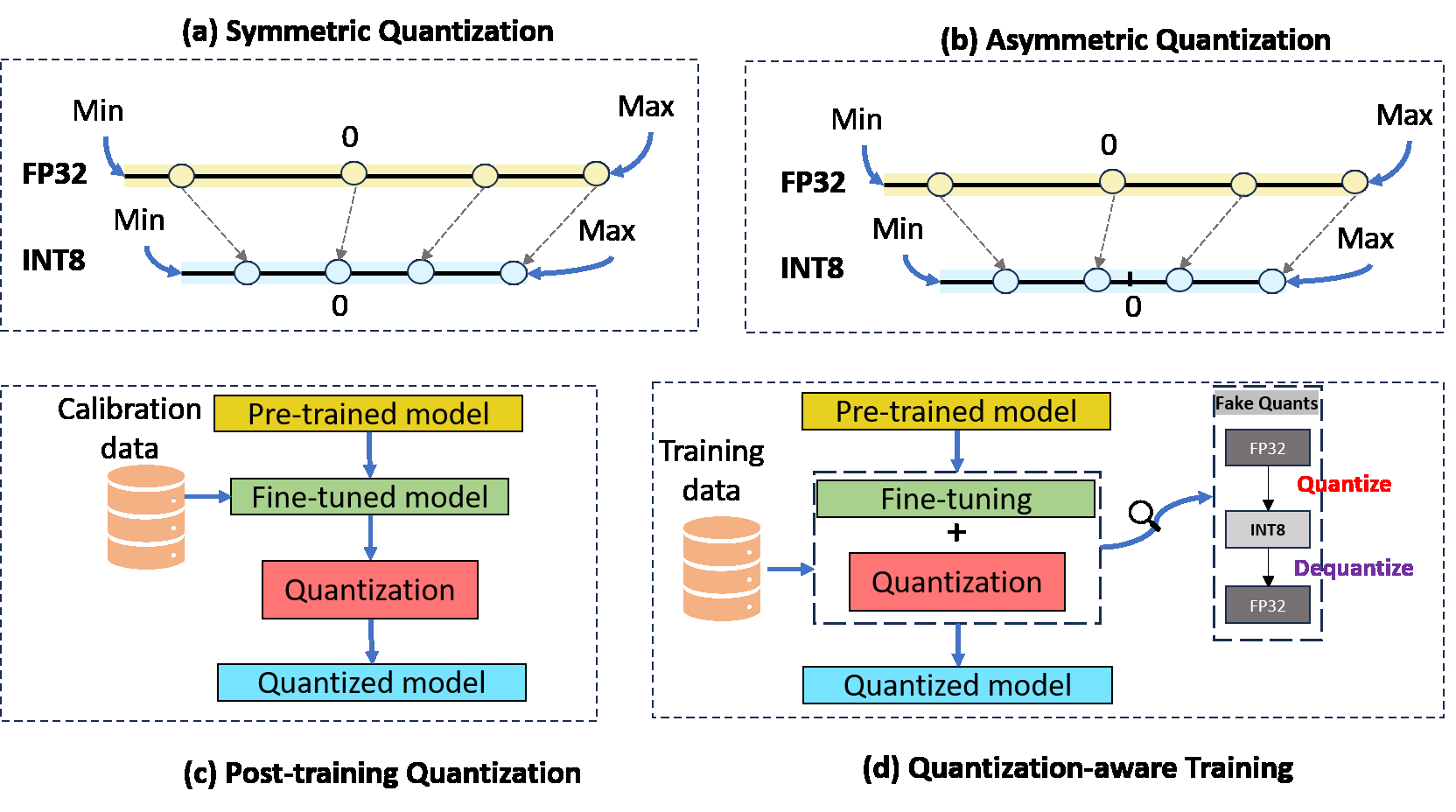}
    \caption{Overview of quantization: symmetric and asymmetric methods alongside post-training quantization (PTQ) and quantization-qware training (QAT) approaches.}
    \label{fig:Quantization}
\end{figure}

Quantization refers to the compression of weights and activations by reducing their precision, which enables computational efficiency. Quantization minimizes memory footprint while maintaining accuracy as it maps continuous values $x$ (e.g., in FP32) to discrete values $q(x)$ in lower-bit formats. In terms of precision, FP32 provides 7 digits of accuracy, FP16 reduces this to 3-4 decimal places, while INT8 eliminates decimals entirely. Despite the lower precision, INT8 quantization is effective because it optimizes the numerical range required for model parameters rather than mapping the entire range of floating-point values. For example, mapping FP32 values $[-3.4 \times 10^{38}, 3.4 \times 10^{38}]$ to the INT8 range $[-128, 127]$ allows efficient representation of the model's parameter range while significantly reducing computational complexity. Quantization improves hardware efficiency as NVIDIA GPUs use tensor cores optimized for faster and cheaper lower-bit operations like INT8 compared to FP32. This delivers high throughput, making it beneficial for resource-limited layers. Quantization methods can be categorized into symmetric and asymmetric approaches. Symmetric quantization maps the range of floating-point values symmetrically around zero. The scale factor $\eta$ is computed as:
\begin{equation}
\eta = \frac{2^{n-1} - 1}{\beta}
\end{equation}
where $\beta$ is the maximum absolute value, and $n$ represents the number of bytes to be quantized. The quantized value $q(v)$ is then calculated as:
\begin{equation}
q(v) = \text{round}(\eta \cdot v)
\end{equation}
This approach ensures that zero remains zero in the quantized space, while maintaining symmetry around the origin. Asymmetric quantization introduces a shift $\gamma$ (zero-point) to handle distributions not centered around zero. The quantized value $q(v)$ is defined as:
\begin{equation}
q(v) = \text{round}(\eta \cdot v + \gamma)
\end{equation}
Here, the scale factor $\eta$ and offset $\gamma$ are calculated as:
\begin{equation}
\eta = \frac{\theta_{\text{max}} - \theta_{\text{min}}}{2^{n-1}}, \quad \gamma = \text{round}(-\eta \cdot \theta_{\text{min}})
\end{equation}
where $\theta_{\text{max}}$ and $\theta_{\text{min}}$ are the maximum and minimum values of the floating point range.
Quantization is broadly classified into Post-Training Quantization (PTQ) and Quantization-Aware Training (QAT). Figure \ref{fig:Quantization} illustrates these approaches.

\subsubsection{Post-Training Quantization (PTQ)}
PTQ updates the weights and activations after training using a small dataset, employing either symmetric or asymmetric methods. It is simpler for weights due to direct access to weight tensors but more challenging for activations as they require measurement with input data. This requires innovative techniques to balance efficiency and accuracy, especially when handling the complexities of activation quantization. \cite{wang2024qvlm} propose Q-VLM, a PTQ framework for large vision-language models (LVLMs) that focuses on minimizing cross-layer dependency errors using activation entropy as a proxy. This method partitions the model into blocks for efficient quantization and employs visual encoder optimization to reduce search space while maintaining accuracy. It achieves significant memory compression and faster inference speeds without performance degradation. \cite{li2024mbq} introduce Modality-Balanced Quantization (MBQ) for LVLMs that accounts for sensitivity differences between vision and language tokens. By balancing reconstruction loss during calibration, MBQ improves accuracy and supports weight-only and weight-activation coquantization, achieving up to 1.4× decoding speedup with a custom W3 kernel. 
NoisyQuant \cite{Liu2023NoisyQuant} reshapes activation distributions using uniform noisy bias to reduce quantization errors in vision transformers and can be adapted to VLMs for efficient activation quantization. To mitigate the performance drop problem in PTQ, P4Q \cite{sun2024p4q} introduces learnable prompts and a lightweight low-bit adapter to realign image and text feature distributions. This approach enhances recognition performance and achieves results comparable to full-precision models while maintaining computational efficiency. Some PTQ methods are specifically designed for vision and language models independently; however, they can also be applied to vision and language components within VLMs. Therefore, it is essential to consider these methods when discussing PTQ strategies for VLMs. \cite{Yuan2022PTQ4ViT} propose twin uniform quantization for vision transformers to handle the special distributions of activations and a Hessian-guided metric to improve scaling factor selection. This can enhance the efficiency of transformer-based vision encoders in VLMs while maintaining near-lossless accuracy. Similarly, \cite{Lv2024PTQ4SAM} develop PTQ4SAM, a quantization framework for the Segment Anything Model (SAM), which incorporates a Bimodal Integration strategy to address challenging activation distributions and Adaptive Granularity Quantization to improve softmax quantization. PTQ techniques such as LRQuant and BiLLM offer promising solutions for compressing language components in VLMs. LRQuant \cite{zhao2024lrquant} enhances learnability and generalization through dynamic smoothing and test-time adaptation, while BiLLM \cite{Huang2025BiLLM} improves and extends ultra-low-bit quantization to minimize memory and computation costs.

\subsubsection{Quantization-Aware Training (QAT)}
\label{QAT}
QAT occurs during training itself, allowing the model to fine-tune with quantized weights and achieve better performance than PTQ. Fake quantization is applied in the forward pass, making the model adapt to quantization effects while retaining the full precision version. Very few studies have directly investigated this approach in the context of VLMs. \cite{Xie2024QSLAW} introduce QSLAW, a quantization-aware scale learning approach for VLMs, optimizing multimodal large language models (MLLMs) by learning group-wise scale factors to mitigate quantization errors. Additionally, it uses a modality-aware warmup approach to enhance VL instruction tuning efficiency while  preserving linguistic knowledge. Most studies have focused on either the vision or language component, but surveying these methods is valuable as they can be integrated. Q-ViT \cite{Li2022QViT} proposes a fully differentiable QAT framework for Vision Transformers, optimizing both quantization scales and bit-widths dynamically. It employs head-wise bit-width allocation and a switchable scale mechanism to enhance quantization robustness, achieving 3-bit precision with minimal accuracy loss. GPUSQ-ViT \cite{Yu2023GPUSQViT} employs sparse distillation-aware QAT to optimize Vision Transformers for GPU efficiency. By integrating feature-based knowledge distillation with quantization scaling, it minimizes accuracy degradation while achieving substantial reductions in model size. LLM-QAT \cite{Liu2023LLMQAT} leverages self-generated data from the pre-trained model for knowledge distillation, enabling QAT without access to the original dataset. It quantizes weights, activations, and the KV cache and significantly improves performance over PTQ methods. \cite{Chen2024EfficientQAT} employs QAT (EfficientQAT) to optimize LLMs by introducing block-wise training of all parameters (Block-AP) and end-to-end training of quantization parameters (E2E-QP). This approach reduces memory overhead and training time while maintaining high accuracy in low-bit quantization scenarios. 

\subsection{Low-rank Approximation}
Low-rank approximation is a mathematical technique that reduces the number of parameters in a model. It discovers latent patterns and eliminates redundancy. The model becomes more efficient as a result of this reduction. The key idea is to shrink large matrices while retaining most of their information (for example, a 5000 × 5000 weight matrix can be approximated using 5000 × 500 and 500 × 5000 matrices, reducing the number of parameters from 25 million to 500 thousand). Matrix "A" can be approximated as follows:
\begin{equation}
A \approx YZ^\top
\end{equation}
where each column of \( A \) is a linear combination of the columns of \( Y \) and each row of \( A \) is a linear combination of the rows of \( Z^\top \).

\begin{figure}
    \centering
    \includegraphics[width=\textwidth]{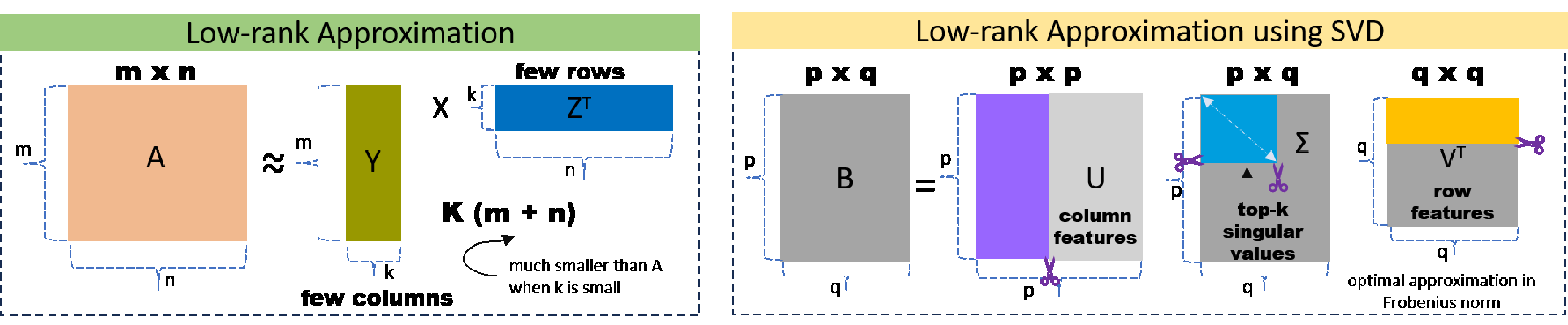}
    \caption{Schematic representation of low-rank approximation and SVD.}
    \label{fig:Low-rank Approximation}
\end{figure}

Low-rank approximation is widely used for image compression \cite{KUMAR2022ImageCompression}, denosing \cite{Guo2019Denoising} and matrix completion \cite{Nie2019MatrixCompletion}. It offers several advantages, such as reduced memory consumption and computational efficiency, making it well suited for deployment on smartphones and IoT devices. It also reduces inference latency, thus enhancing real-time performance. However, the choice of rank is crucial when deploying on resource-constrained devices, as it directly impacts the compression ratio and the overall model performance. The rank-$k$ approximation of a matrix $B$ is computed by decomposing it into its singular value components and retaining only the $k$ most significant ones (singular value decomposition (SVD) \cite{Klema1980SVD} is one such optimal method \ref{fig:Low-rank Approximation}):
\begin{equation}
B_k = \sum_{i=1}^{k} s_i u_i v_i^T
\end{equation}
where $s_i$ are the singular values of $B$, sorted in descending order, and $u_i$ and $v_i$ are the corresponding left and right singular vectors.

Low-rank approximation has been widely used for language models and has recently been applied to multimodal models as well. \cite{Song2024LoRA-Sparse} introduce LoRA-Sparse, a low-rank approximation method for sparse attention in LLMs and multimodal models like LLaVA. LoRA-Sparse employs order-mimic training to approximate full attention effectively, demonstrating benefits across NLP and multimodal benchmarks. SeTAR \cite{Yixia2024SeTAR} proposes a training-free selective low-rank approximation method for OOD detection in CLIP-based models, enhancing efficiency and accuracy without requiring fine-tuning. SeTAR+FT \cite{Yixia2024SeTAR} extends this by integrating fine-tuning for further performance gains, achieving state-of-the-art results on ImageNet1K and Pascal-VOC benchmarks. PELA \cite{Guo2024PELA} incorporates an intermediate pre-training stage with low-rank approximation to compress pre-trained models. It enables efficient fine-tuning by distilling features and regularizing weight perturbations to improve scalability for downstream tasks.

\subsection{Pruning}
\begin{wrapfigure}{l}{0.25\linewidth}
    \centering
        \includegraphics[width=0.25\textwidth]{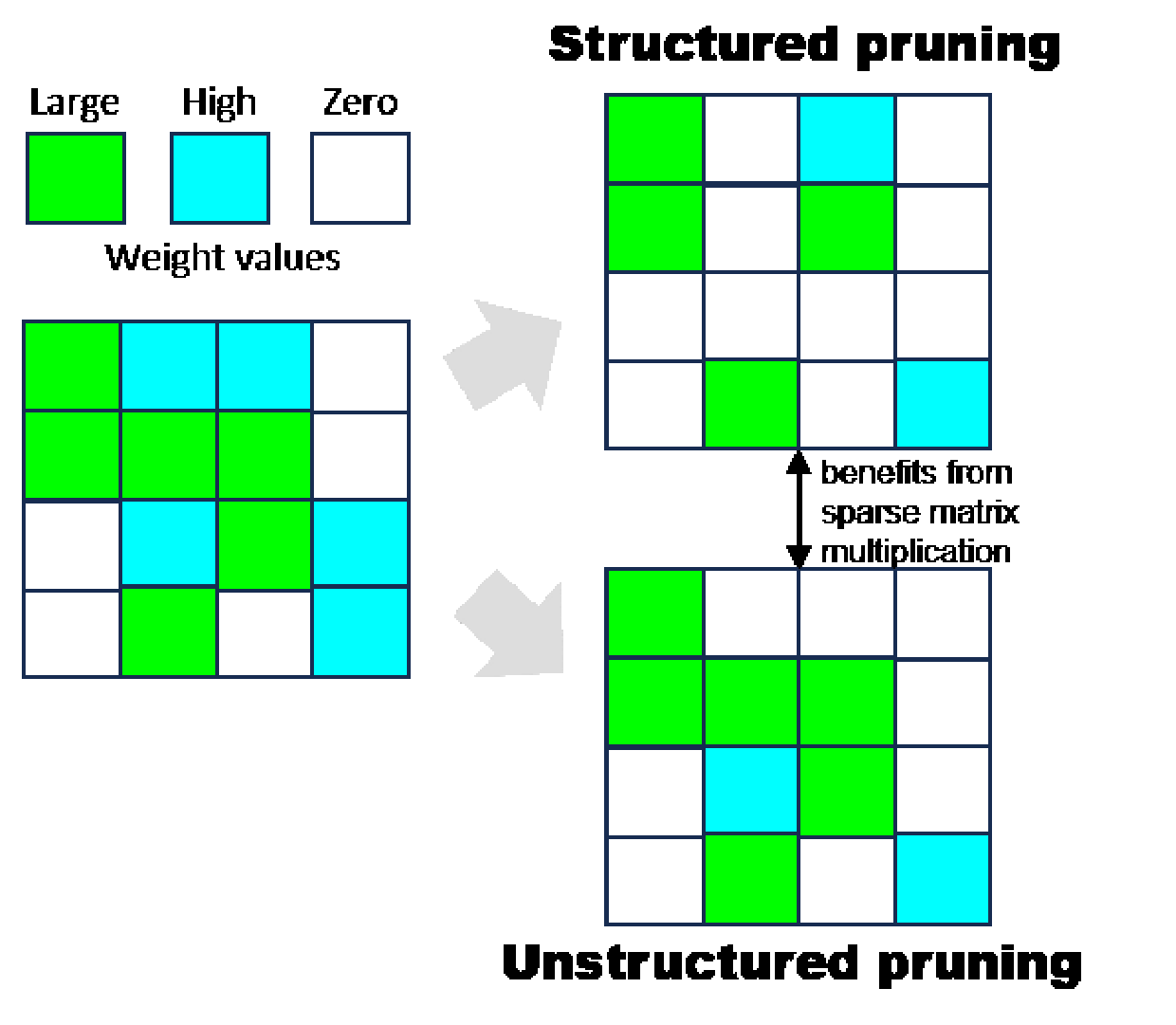}
        \caption{Illustration of structured and unstructured pruning.}
        \label{fig:parameter_pruning}
\end{wrapfigure} Parameter pruning involves removing redundant or less significant model weights. A simple yet effective pruning method is to remove model weights that fall below a predetermined threshold. However, more advanced techniques exist, such as gradient-based pruning, which evaluates weight importance based on sensitivity to loss function, and Hessian-based pruning that uses second-order information to identify less impactful weights. Parameter pruning reduces the memory footprint and computational load of large networks, making them more efficient for resource-constrained devices. By eliminating redundant weights, pruning enables faster inference, lower power consumption, and efficient deployment on edge devices. Pruning is categorized into Structured and Unstructured Pruning. A schematic representation for the same is shown in figure \ref{fig:parameter_pruning}.

\subsubsection{Structured Pruning}
This pruning technique uses a structured strategy to target sub-blocks, rows, or columns. This reduces computational complexity during the forward pass of the weight graph. As a result, hardware utilization becomes more efficient. For example, \cite{Lin2024MoPE-CLIP} introduce MoPE-CLIP, a mask-free structured pruning framework for VLMs, using the Module-Wise Pruning Error (MoPE) metric to assess module importance for cross-modal tasks. It optimizes model compression while preserving performance through width to depth pruning and knowledge distillation. In order to improve the scalability of structured pruning in VLMs and remove the need for costly fine-tuning, \cite{Meng2024OSSCAR} develop OSSCAR, a combinatorial optimization framework that uses low-rank updates and layer-wise reconstruction for efficient local search. Existing Transformer-based VLMs suffer from computational inefficiencies due to redundant token representations and attention heads, leading to high inference costs. To address this \cite{Wang2023SmartTrim} propose SmartTrim an adaptive pruning framework that dynamically trims tokens and attention heads based on cross-modal complexity, ensuring efficient computation while maintaining performance across diverse tasks. Similarly, Transformer models consist of heterogeneous sub-structures making global pruning unreliable. To tackle this, \cite{fang2024isomorphic} introduce Isomorphic Pruning, which groups isomorphic sub-structures based on computational topology. This leads to reliable pruning.

\subsubsection{Unstructured Pruning}
In this method, individual model weights are pruned. Unstructured pruning offers more flexibility compared to its structured counterpart and experiences a lower accuracy drop. However, this causes non-zero weights to be distributed unevenly, which results in irregular sparsity. Edge hardware is optimized for dense matrix operations, but unpredictable non-zero weight values can lead to frequent cache misses. To address this, various sparsification methods are being investigated and hardware accelerators are being improved. Some studies have explored unstructured pruning for VLMs. For example, \cite{Sung2024ECoFLaP} introduces a two-stage coarse-to-fine layer-wise unstructured pruning approach that first determines sparsity ratios using a global importance score and then performs local pruning. This approach uses zeroth-order gradient approximations to reduce computational overhead while maintaining model performance across a variety of multimodal datasets. To solve the challenge of task-agnostic pruning and preserving multimodal transferability, MULTIFLOW \cite{farina2024multiflow} presents a gradient-free pruning framework that ranks parameter importance based on both magnitude and information flow. Also, \cite{he2024rethinkingpruningvisionlanguagemodels} reveal that unstructured pruning in VLMs significantly degrades performance beyond 50\% sparsity.

\subsection{Knowledge Distillation (KD)}
\begin{figure}[htbp]
    \centering
    \includegraphics[width=\textwidth]{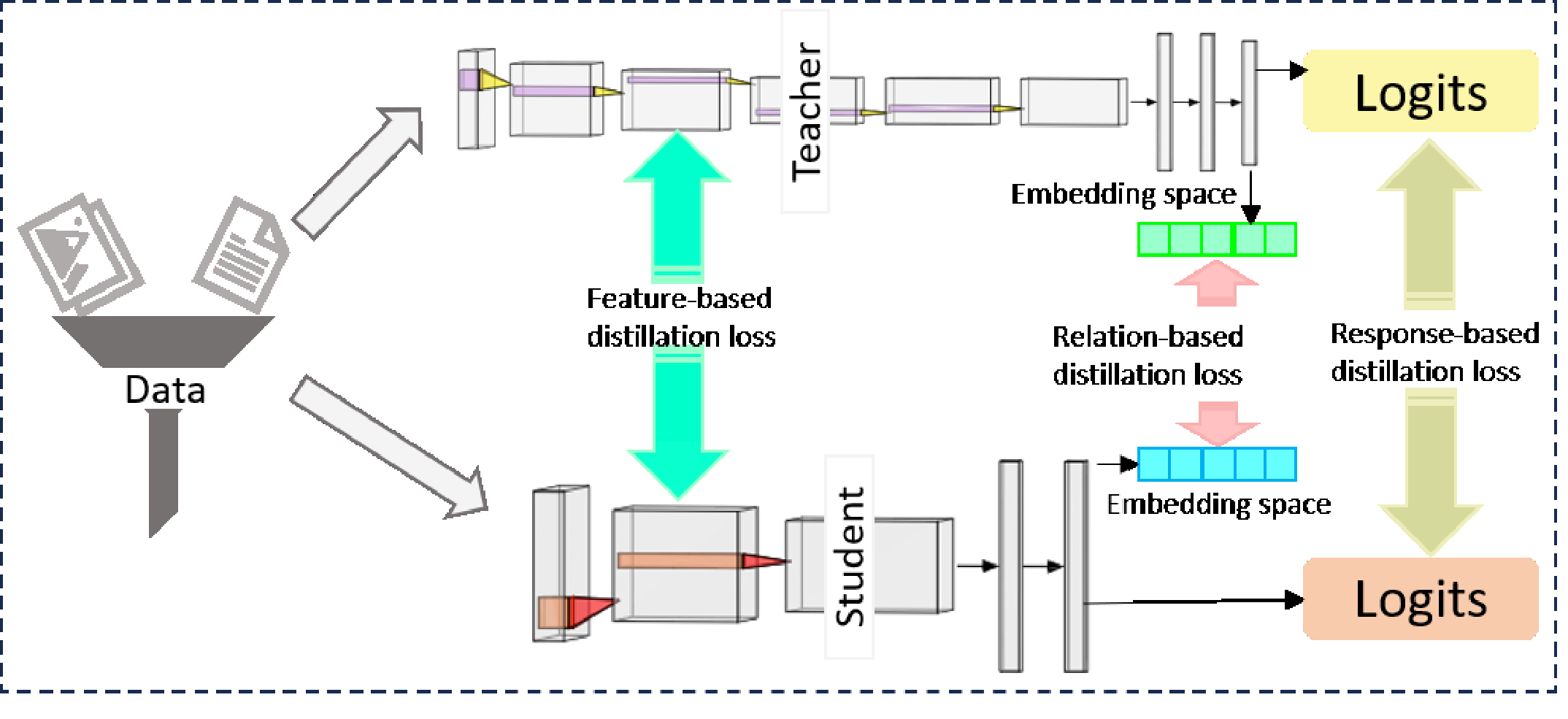}
    \caption{Diagrammatic representation of \textbf{1.} response-based, \textbf{2.} feature-based, and \textbf{3.} relation-based knowledge distillation. In \textbf{1.}, the student mimics the teacher’s soft labels. In \textbf{2.}, the student replicates the teacher’s feature activations. In \textbf{3.}, the student preserves the relational structure of the teacher’s data embeddings.}
    \label{fig:Knowledge Distillation}
\end{figure}

KD is a model compression technique in which a teacher network transfers knowledge to a smaller student network. The teacher network is typically a large neural network with many parameters. Knowledge can reside in internal layers, activations, or even soft labels \ref{fig:Knowledge Distillation}. KD enables deployment of a smaller high-performance model on edge devices. Several studies have used KD in VLMs to enhance efficiency. \cite{Li2024PromptKD} introduce PromptKD, a novel approach that distills knowledge from a pretrained CLIP teacher model to a lightweight student using prompt-based imitation, enabling effective adaptation to domain-specific tasks without requiring labeled data. PromptKD additionally allows fast inference by leveraging pre-stored text features. VLMs still struggle with understanding spatial and contextual relationships. To address this  SF-CLIP \cite{Sameni2024SF-CLIP} employ masked knowledge distillation from pre-trained vision and language models, ensuring better spatial feature learning and multilingual retrieval while preserving training efficiency. \cite{Fang2021KD} propose a distillation approach for compressing large visual-linguistic models by aligning student and teacher model representations. They ensure consistency in visual tokens using the same region proposals and introduce loss functions to match attention distributions and hidden states, improving efficiency without sacrificing performance. \cite{liu2022KD-VLP} develop KD-VLP, an object-aware framework that embeds object-level knowledge into multimodal learning. By leveraging knowledge distillation from object features and semantic labels, their method improves cross-modal alignment and enhances downstream task performance without requiring external object detectors during inference. VLM-based detectors struggle with domain adaptation due to reliance on biased object proposals from the target domain, making direct knowledge distillation ineffective. To solve this issue, VLDadaptor \cite{Ke2024VLDadaptor} leverage domain-mixed contrastive knowledge distillation and domain-mixed consistency distillation to align category-level features in the visual-language embedding space. Even the recent breakthrough, DeepSeek-R1 \cite{deepseekR1}, demonstrates that distilling its outputs to a student model (DeepSeek-R1-7B) surpasses previous non-reasoning models.

\vspace{1pt}

\subsection{Other Methods} 
\begin{wrapfigure}{l}{0.35\linewidth}
    \centering
        \includegraphics[width=0.35\textwidth]{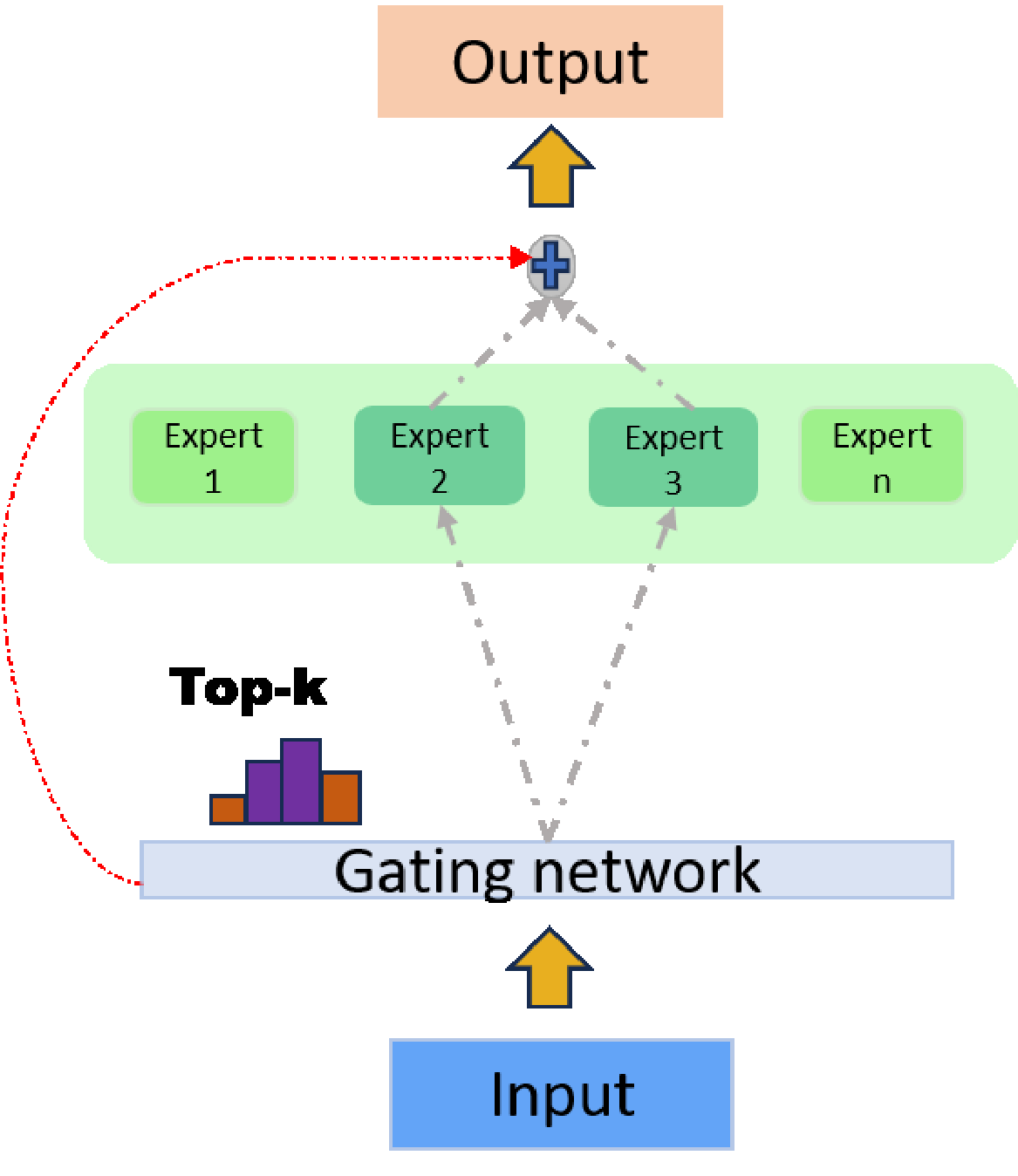}
        \caption{Functioning of mixture of experts (MoE) with a gating mechanism.}
        \label{fig:MoE}
\end{wrapfigure} Recent pre-deployment work has focused on developing mixture of experts (MoE) and adaptive attention mechanisms for resource-constrained devices. Mixture of Experts enhances model efficiency by dynamically selecting specialized subnetworks or "experts," for different inputs. MoE facilitates accelerated inference relative to a model with the same number of parameters. Figure \ref{fig:MoE} shows the functioning of MoE with a gating mechanism to pass information to multiple experts. Load balancing is essential during training to avoid bias toward specific experts \cite{ProphetLoadBalancing}. Recent research leverages MoE to enhance efficiency in VLMs.  \cite{Yu2024BoostingCL} introduce a parameter-efficient continuous learning framework for VLMs by integrating MoE adapters. MoE facilitates zero-shot capabilities through the dynamic selection of task-specific adapters. \cite{shen2023scalingMoE} illustrate the importance of MoE in scaling VLMs while balancing the trade-off between computational resources and performance. Med-MoE \cite{jiang2024MedMoE} leverages domain-specific experts and a meta-expert to efficiently handle generative and discriminative medical tasks while reducing the number of parameters during inference. Another research focus is optimizing attention mechanisms to maintain performance while reducing FLOPs. Adaptive attention \cite{Lu2016AdaptiveAttention} has emerged as a promising approach that  selectively attends to tokens based on context, as compared to previous mechanisms that attended to all tokens \cite{Vaswani2017Attentionisallyouneed}.

\section{Efficient Fine-tuning}
\label{efficientfinetuning}
Efficient fine-tuning refers to optimizing pre-trained models by updating only a small subset of parameters. Fine-tuning eliminates the need for full retraining, thereby reducing memory consumption. Fine-tuning techniques include both parameter-efficient and memory-efficient methods.

\subsection{Parameter-Efficient Fine-Tuning (PEFT)}
PEFT is a computationally efficient method for adapting neural networks by introducing a small set of trainable parameters in the last layer while keeping the remaining model's parameters and architecture unchanged. This differs from standard fine-tuning, which updates all parameters in the network. PEFT offers the following advantages: \blackcircled{1} it reduces model size when compared to standard fine -tuning, \blackcircled{2} mitigates the forgetting phenomenon, \blackcircled{3} and improves generalization to unseen contexts. PEFT includes various methods such as low-rank adaptation (LoRA), prompt tuning, adapter-based, and mapping-based approaches.

\subsubsection{Low-Rank Adaptation (LoRA)}
\begin{wrapfigure}{l}{0.35\linewidth}
    \centering
        \includegraphics[width=0.35\textwidth]{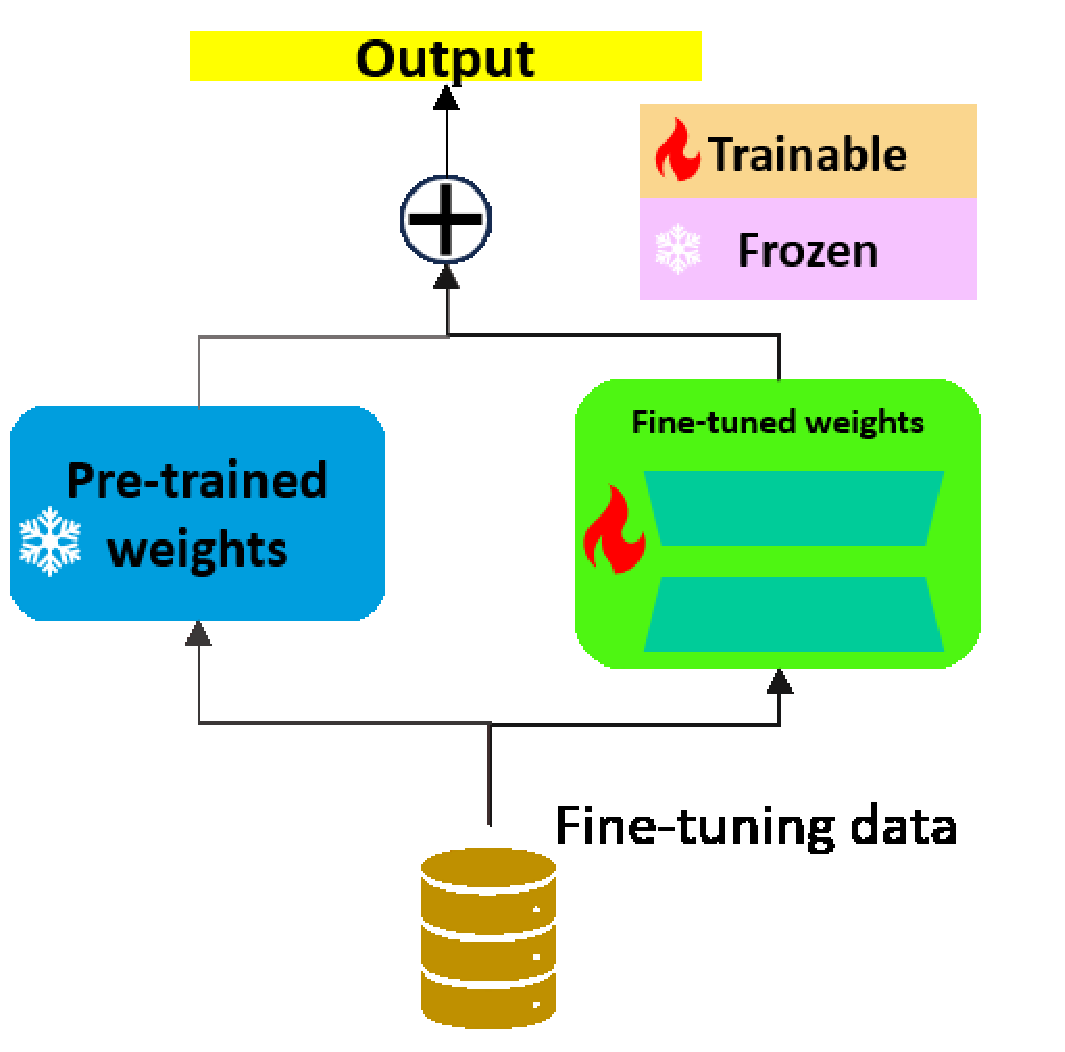}
        \caption{Illustration of LoRA.}
        \label{fig:LoRA}
\end{wrapfigure}

Standard fine-tuning requires updating all model parameters, which is not only time-intensive but also requires substantial storage. In contrast, LoRA uses low-rank decomposition to approximate the weight update matrix (refer figure \ref{fig:LoRA}). Instead of altering the entire weight matrix  \( W \in \mathbb{R}^{m \times n} \), LoRA parameterizes the update matrix $\Delta {W}$ as the product of two lower-dimensional matrices, capturing essential task-specific adaptations while minimizing storage and computational costs.
LoRA decomposes the update matrix \( \Delta W \) as:
\begin{equation}
\Delta W = P Q^{T}, \quad \text{where } P \in \mathbb{R}^{m \times r}, \quad Q \in \mathbb{R}^{n \times r}
\end{equation}
The updated weight matrix is given by:
\begin{equation}
W^* = W + \Delta W = W + P Q^{T}
\end{equation}
 With VLMs gaining significant attention, some studies have explored applying LoRA to enhance their efficiency. \cite{bai2024adapterensemble} propose an adapter ensemble strategy to efficiently fine-tune large-scale VLMs while integrating LoRA to mitigate the additional parameter burden introduced by the ensemble process. By including LoRA, they aim to reduce memory and compute overhead while maintaining competitive performance, particularly in vision-language retrieval tasks. CLIP-LoRA \cite{Zanella2024LowRankFA} reduces training overhead while maintaining strong performance across 11 datasets, demonstrating that LoRA can outperform prompt tuning and adapter-based approaches in few-shot scenarios without requiring extensive hyperparameter tuning. \cite{Yin2023Dense, Agiza2024MTLoRA, Zhu2024MeLo} have focused on vision models, while \cite{Jin2024DerivativeFree, Yu2023Speech} have explored language models; however, limited research has been conducted on multimodal settings like VLMs.

\subsubsection{Prompt Tuning}
\begin{wrapfigure}{l}{0.35\linewidth}
    \centering
        \includegraphics[width=0.35\textwidth]{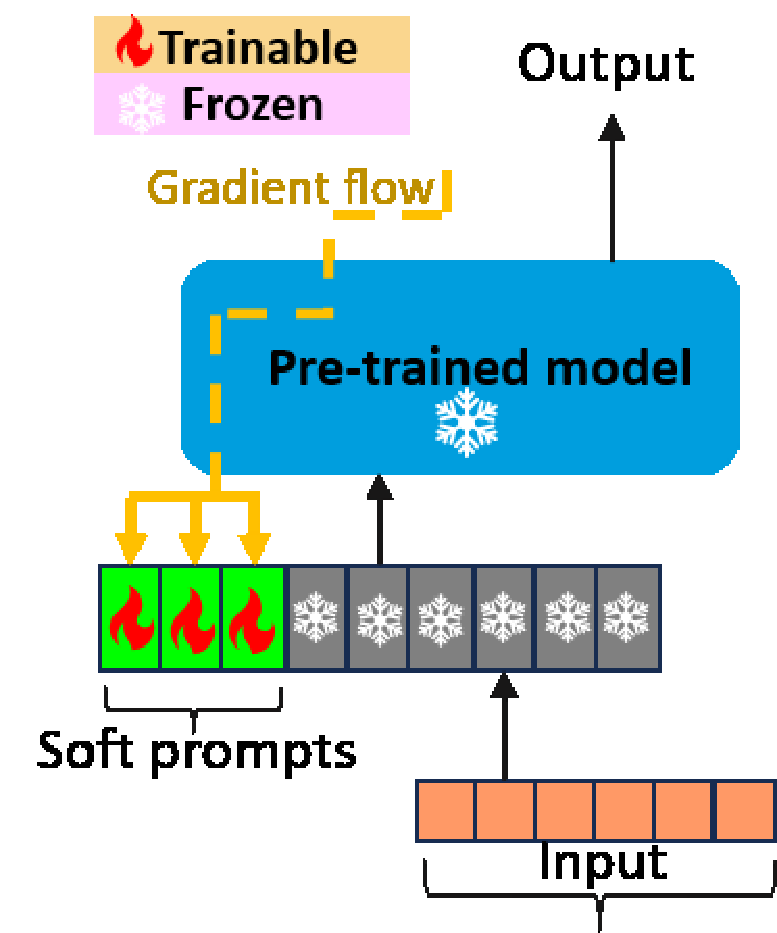}
        \caption{Illustration of prompt tuning.}
        \label{fig:Prompt Tuning}
\end{wrapfigure}

In prompt tuning \cite{lester2021PowerPT}, task-specific "soft prompts" are added to the input data. These are opaque learnable tensors concatenated with the input embedding as shown in figure \ref{fig:Prompt Tuning}. Throughout training, only the prompts are modified for the specified task. This strategy is extremely beneficial for multitasking and continual learning.
 \cite{Yao2023KgCoOp} introduce Knowledge-guided Context Optimization (KgCoOp) to enhance the generalization capability of prompt tuning for VLMs. Their strategy reduces the discrepancy between learned soft prompts and hard prompts, thus reducing the likelihood of forgetting important general knowledge. By integrating this optimization constraint with contrastive learning, their method improves performance on unseen classes while also maintaining computational efficiency. \cite{Xing2024DPT} propose a dual-modality prompt tuning (DPT) approach that optimizes both visual and textual prompts simultaneously to adapt vision-language models to downstream tasks. They also introduce class -aware visual prompts, generated by cross-attention between textual features and image tokens, to improve task-specific feature extraction. To address the misalignment between text and image embeddings in VLMs, \cite{Cho2023DAPT} develop DAPT to optimize feature distributions by maximizing inter-class separation and minimizing intra-class variability. This approach improves generalization in few-shot learning and domain adaptation.

\subsubsection{Adapter-based Methods}
\FloatBarrier

\begin{wrapfigure}{l}{0.45\linewidth}
    \centering
    \vspace{-5pt} 
    \includegraphics[width=0.43\textwidth]{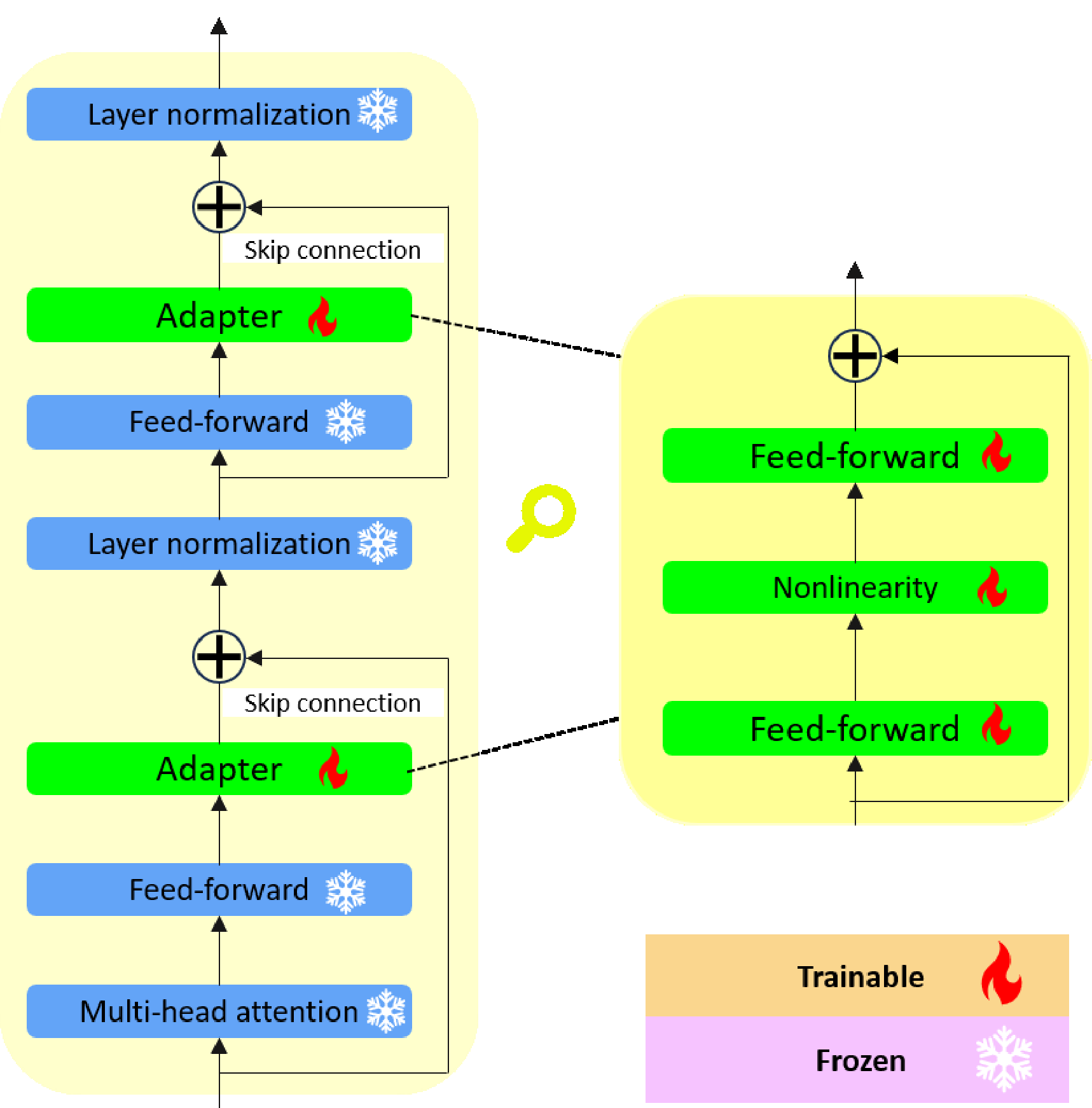}
    \vspace{-5pt} 
    \caption{Illustration of adapters in a transformer block.}
    \label{fig:Adapters}
    \vspace{40pt} 
\end{wrapfigure}

Adapter-based methods \cite{Houlsby2019ParameterEfficientTLAdapter} introduce randomly initialized bottleneck layers "$A$" within transformer layers, as illustrated in Figure~\ref{fig:Adapters}. During training, adapters learn a transformation function $f_A$ that modifies the activations of the model "$h$".
\begin{equation}
h' = f_A(h)
\end{equation}
Adapters consist of feed-forward layers, where one layer projects the input to a lower-dimensional representation, and the other restores it to the original space. This transformation can be expressed as:
\begin{equation}
X \in \mathbb{R}^{a \times a} \rightarrow W_d X \in \mathbb{R}^{a \times b} \rightarrow W_u (W_d X) \in \mathbb{R}^{a \times a}
\end{equation}
where \( W_d \in \mathbb{R}^{b \times a} \) reduces the dimension and \( W_u \in \mathbb{R}^{a \times b} \) restores it. Adapters have been extensively used in large language models (LLMs) and have recently gained traction in VLMs.
\cite{Sung2022VL-ADAPTER} propose adapter-based parameter-efficient transfer learning techniques for vision and language tasks. They evaluate adapter variants in a unified multi-task setup across image-text and video-text benchmarks, demonstrating that weight-sharing adapters achieve competitive performance with only 3.39-4.18\% of updated parameters. \cite{Yang2024MMA} introduce MMA (Multi-Modal Adapter) to enhance alignment between vision and language representations in VLMs by selectively integrating adapters into higher transformer layers to balance discrimination and generalization. Their approach achieves state-of-the-art performance across novel class generalization, cross-dataset adaption, and domain generalization tasks. \cite{cHENG202Meta-adapter} develop Meta-Adapter, a residual-style adapter that enables online few-shot learning for CLIP by refining features with few-shot samples to enhance generalization.

\subsubsection{Prefix Tuning}
\begin{wrapfigure}{l}{0.35\linewidth}
    \centering
        \includegraphics[width=0.35\textwidth]{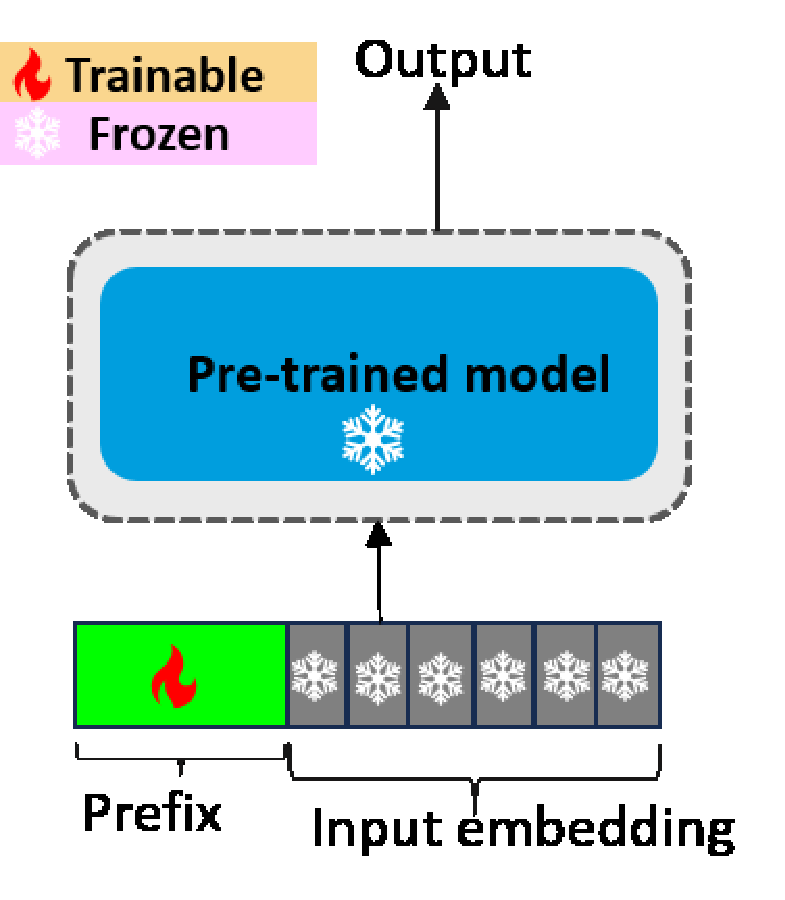}
        \caption{Illustration of prefix tuning.}
        \label{fig:Prefix Tuning}
\end{wrapfigure} Prefix Tuning \cite{li2021prefixtuning} introduces task-specific continuous vectors into each layer of the transformer architecture. Prefixes are abstract representations concatenated with input embeddings in each layer. \cite{MedVQA2023} introduce a mapping network that transforms visual features into token embeddings and employs prefix tuning to to refine query  representations in each transformer layer for open-ended medical visual question answering (VQA). \cite{song2024contextaware} propose a vision-language storytelling framework that integrates a trainable transformer mapping network and a multimodal contrastive objective to enhance coherence in generated narratives. Prefix tuning is employed to align CLIP-extracted visual features with a frozen GPT-2 \cite{Radford2019gpt-2} model, facilitating efficient vision-language alignment. Prefixes are introduced in personalized image captioning to map visual features with user context, enabling a frozen GPT-2 model to provide personalized captions \cite{Wang2023User-Aware_Prefix-Tuning}.

\subsection{Memory-Efficient Fine-Tuning (MEFT)}
Memory-Efficient Fine-Tuning (MEFT) \cite{Liao2023MEFT} optimizes large pre-trained models for downstream tasks while minimizing memory overhead. Unlike traditional fine-tuning, which involves backpropagation through the entire model, MEFT reduces activation storage and memory-intensive gradient computations. One such method, LoSA \cite{Mercea2024TimeMA} avoids backpropagation through the backbone. This is achieved by incorporating a lightweight parallel network that refines features from the backbone. Transformer models remain integral to VLMs, making MEFT-based optimization crucial for efficient fine-tuning across diverse applications. One such method is SLIMFIT \cite{ardakani-etal-2024-slimfit}, which dynamically freezes less-contributory layers based on training dynamics and combines this with quantization and pruning to reduce activation memory. This method saves up to 3.1× on-device memory and maintains accuracy within 0.4\% of full fine-tuning. \cite{liu2024m2ist} develop \(M^2IST\) (Multi-Modal Interactive Side-Tuning) to enhance vision-language alignment and enable memory-efficient fine-tuning (MEFT) for referring expression comprehension (REC). By freezing pre-trained encoders and updating lightweight side networks, \(M^2IST\) reduces GPU memory usage by 60.39\% compared to full fine-tuning.

\section{Runtime Optimizations}
\label{runtime optimizations}
Runtime optimization refers to techniques that dynamically adjust computations during  inference to improve computational efficiency, memory use, and energy consumption without requiring retraining. Unlike static optimizations such as pruning or quantization, runtime methods adapt model execution on-the-fly based on input complexity, hardware constraints, or task-specific requirements. VLMs are computationally intensive, necessitating significant processing for multimodal interactions. Their deployment on  resource-constrained devices is challenging due to the high memory, power, and latency demands. Runtime optimization mitigates these issues by reducing unnecessary computations and accelerating inference.

\subsection{Token Reduction}
A major bottleneck in VLMs is redundant token processing, particularly in transformer-based models where  both visual and textual tokens contribute to high computational costs. Since VLMs compute relationships between all visual tokens (\(N_v\)) and textual tokens (\(N_t\)) using cross-attention mechanisms, the computational complexity grows quadratically as $\mathcal{O}(N_v \times N_t)$, leading to significant inefficiencies during multi-modal fusion. Token reduction dynamically prunes, merges, or selectively processes tokens at inference time to optimize speed and memory usage. PuMer dynamically removes redundant tokens and merges similar ones in both vision and language modalities during inference \cite{Cao2023PuMerPA}. Unlike pre-deployment pruning, PuMer adapts token selection per instance to improve efficiency while maintaining performance. Turbo \cite{Ju2024Turbo} employs mutual redundancy and semantic value-based token pruning, using an informativity score to rank tokens. This score evaluates redundancy and contribution to semantics, enabling Turbo to delete less informative tokens while retaining crucial ones. This method enhances inference speed and reduces computational overhead in VLMs.

\subsection{Test-Time Adaptation (TTA)}
TTA dynamically adjusts inference strategies in real-time without modifying model weights, allowing models to adapt to domain shifts and unseen data. Unlike traditional fine-tuning, TTA operates on-the-fly during inference. Standard VLMs process inputs using a fixed learned distribution $p(y \mid x; \theta)$, but in real-world settings, the input distribution $p(x)$ often shifts. This causes performance degradation. 
TTA mitigates this by estimating an updated conditional probability:
\begin{equation}
p(y \mid x) = \arg\max_{y} \mathbb{E}_{q(\theta)} [ p(y \mid x; \theta) ]
\end{equation}
where $q(\theta)$ represents an adapted inference distribution that dynamically adjusts based on observed data. \cite{MA2023SwapPrompt} introduce SwapPrompt, a contrastive learning-based approach that dynamically adjusts prompts during inference to improve zero-shot classification by aligning textual representations with appropriate visual features. \cite{fuchs2025onlinegaussiantesttimeadaptation} develop OGA (Online Gaussian Adaptation), a TTA method for VLMs that models visual feature likelihoods using multivariate Gaussian distributions. By incorporating zero-shot priors into a MAP (Maximum A Posteriori) framework, OGA improves robustness without the need  for dataset-specific hyperparameter tuning. TDA (Training-free Dynamic Adapter) \cite{Karmanov2024EfficientTTA} maintains a key-value cache for pseudo labels and test sample features, enabling progressive adaptation without backpropagation. This significantly reduces computational overhead. \cite{Farina2024FrustratinglyEasy} introduce ZERO, a test-time adaptation method that improves VLM robustness without optimization. It marginalizes over augmented views and applies a zero-temperature Softmax, achieving state-of-the-art results while being 10× faster and 13× more memory efficient than test-time prompt tuning. We provide details on TTA techniques in \ref{TTA} and \ref{TPT}.

\subsubsection{Test-Time Augmentation}
\label{TTA}
Test-time augmentation involves applying various transformations to test data and aggregating the model's predictions to enhance performance during inference. By leveraging multiple augmented versions, test-time augmentation improves accuracy without the need for retraining, making it well-suited for low-power inference scenarios. MTA (MeanShift for Test-time Augmentation) \cite{Zanella2024MTA} enhances zero-shot generalization of VLMs by leveraging multiple augmented views without requiring prompt tuning or model retraining. It filters augmented views using inlierness assessment and refines embeddings with robust MeanShift optimization. MTA is $\approx$ 3× faster than the prompt tuning method in computation time. Unlike standard test-time augmentation, \cite{Kimura2024TestTimeAM} use variational Bayes to weight augmentations, suppressing ineffective ones for better predictions. CutMixOut \cite{Fawakherji2024WACVTTA} introduces a novel test-time text augmentation approach for multimodal person re-identification. It applies CutMix and CutOut to text descriptions during inference to improve retrieval accuracy without the need for additional training.

\subsubsection{Test-Time Prompt Tuning (TPT)}
\label{TPT}
TPT optimizes prompts at inference time to enhance the generalization of frozen VLMs.  Given a test sample $x_t$, the goal is to adjust the prompt representation $p_t$ such that the similarity between the test image feature $f_v(x_t)$ and the optimized text feature $f_t(p_t)$ is maximized:
\begin{equation}
p_t^* = \arg\max_{p_t} \; \cos \big( f_v(x_t), f_t(p_t) \big)
\end{equation}
where $f_v(x_t)$ is the visual feature extracted by the vision encoder and $f_t(p_t)$ is the text feature from the language encoder. TPT \cite{shu2022tpt} optimizes prompts using entropy minimization across augmented views  , enhancing zero-shot top-1 accuracy of CLIP by an average of 3.6\%. C-TPT \cite{yoon2024ctpt} improves TPT by using text feature dispersion to enhance calibration while preserving accuracy, achieving better uncertainty quantification without labeled data. Self-TPT \cite{EfficientTPT} further improves TPT with self-supervised learning, achieving strong performance with 25× faster inference and 30× lower memory use.

\section{Privacy-preserving Distributed VLM}
Privacy-preserving distributed VLM ensures that model training occurs across multiple devices or servers without sharing raw data. The key benefits include enhanced data security and a decreased risk of data leakage. Below, we discuss key techniques such as Federated Learning (FL) \ref{sec:FL} and Split Learning \ref{sec:SL}.
\label{privacyprservingdistributedvlm}
\begin{figure}[htbp]
    \centering
    \includegraphics[width=\textwidth]{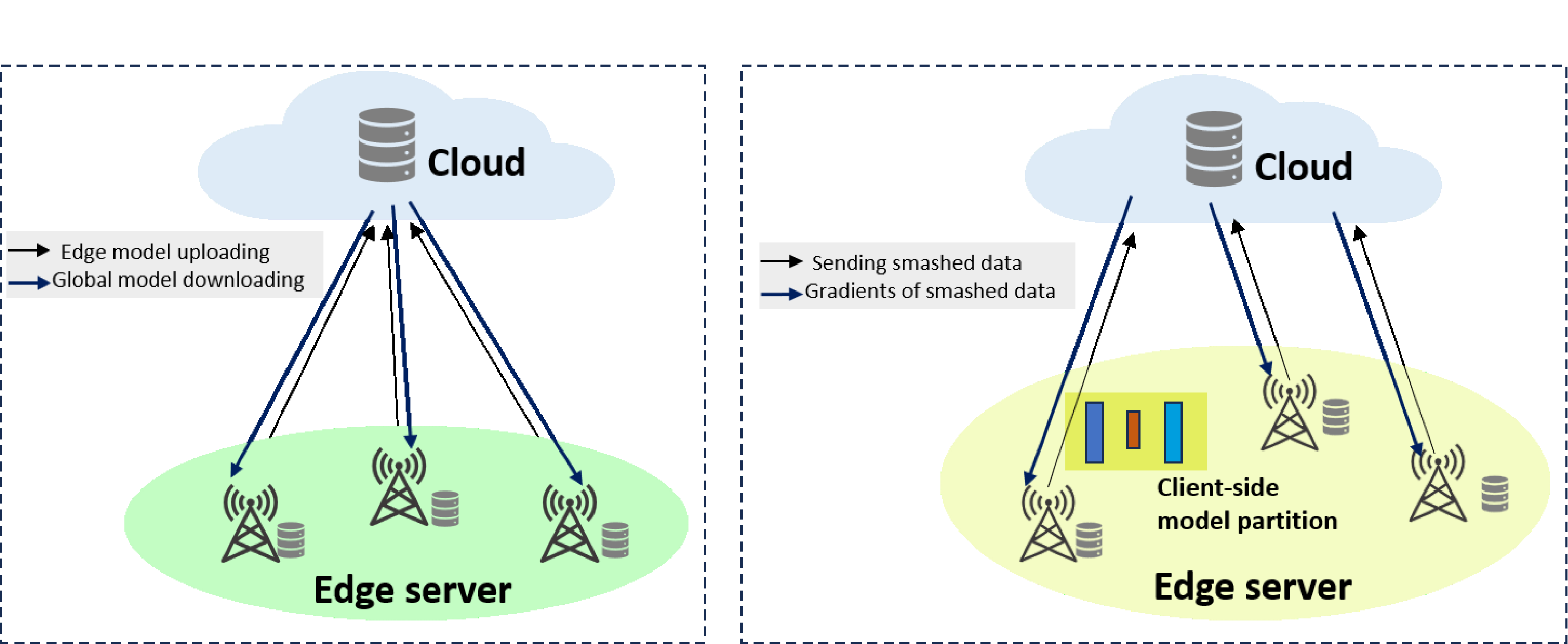}
    \caption{(left) Working of federated learning and (right) split learning.}
    \label{fig:Privacy-preserving}
\end{figure}
\subsection{Federated Learning (FL)}
\label{sec:FL}
FL (figure \ref{fig:Privacy-preserving} (left)) allows multiple clients <$C_1, C_2, \dots, C_N$> to collaboratively update a global model $w$ without sharing raw data. Each client updates $w$ by minimizing a local loss function $L_i$ using its data $D_i$, and sends only model updates $\Delta w_i$ to a central aggregator. The global model (usually a server model) is then updated using techniques such as FedAvg \cite{McMahan2016FedAvg} and FedProx \cite{FedProxUnreliable}. Recently, several studies have explored FL with VLMs \cite{Pan2024TheoreticalAnalysis, zeng-etal-2024-fair}. PMG-FL (personalized and multi-granularity federated learning) \cite{Gao2024PMG-FL} efficiently adapts vision-language foundation models on edge devices, addressing challenges such as data heterogeneity and limited resource availability through  prompt-based local adaptation and knowledge fusion across different device granularities. \cite{yan2024lightweightunsupervisedfederatedlearning} introduce a lightweight unsupervised federated learning framework (FST-CBDG) that leverages pre-trained VLMs to enhance model adaptation on edge devices without requiring labeled data. By refining pseudo-labels through self-training and addressing data heterogeneity with class-balanced data generation, this approach significantly reduces computational and communication overhead. pFedPrompt \cite{Gu02023pFedPrompt} improves personalization in FL by adapting the prompts to user-specific data and using multimodal learning for efficient model alignment.

\subsection{Split Learning}
\label{sec:SL}
Split Learning (figure \ref{fig:Privacy-preserving} (right)) is a distributed learning technique where a deep model is split between clients and a central server, enabling collaborative training without sharing raw data. Split Learning ensures privacy by partitioning the model into client-side $f_c(\theta_c)$ and server-side $f_s(\theta_s)$, where $\theta_c$ and $\theta_s$ represent the parameters of the client and server models, respectively. Instead of transmitting raw data $x$, the client sends only intermediate activations $h = f_c(x; \theta_c)$ to the server. The server then processes $h$ using $f_s(\theta_s)$, ensuring that sensitive data remains local while enabling collaborative model optimization. Although direct applications of split learning to VLMs are scarce, related works provide useful insights. For example, SplitLoRA \cite{lin2024splitlorasplitparameterefficientfinetuning} has been explored for large language models (LLMs), which often serve as the language encoder in VLMs. Similarly, Bidirectional Contrastive Split Learning \cite{Sun2024ContrastiveSL} has been applied for a VQA task, demonstrating split learning's potential in multimodal settings.

\section{Lightweight Vision-Language Models}
\label{Efficient VLMs}
\begin{figure}[htbp]
    \centering
    \includegraphics[width=\textwidth]{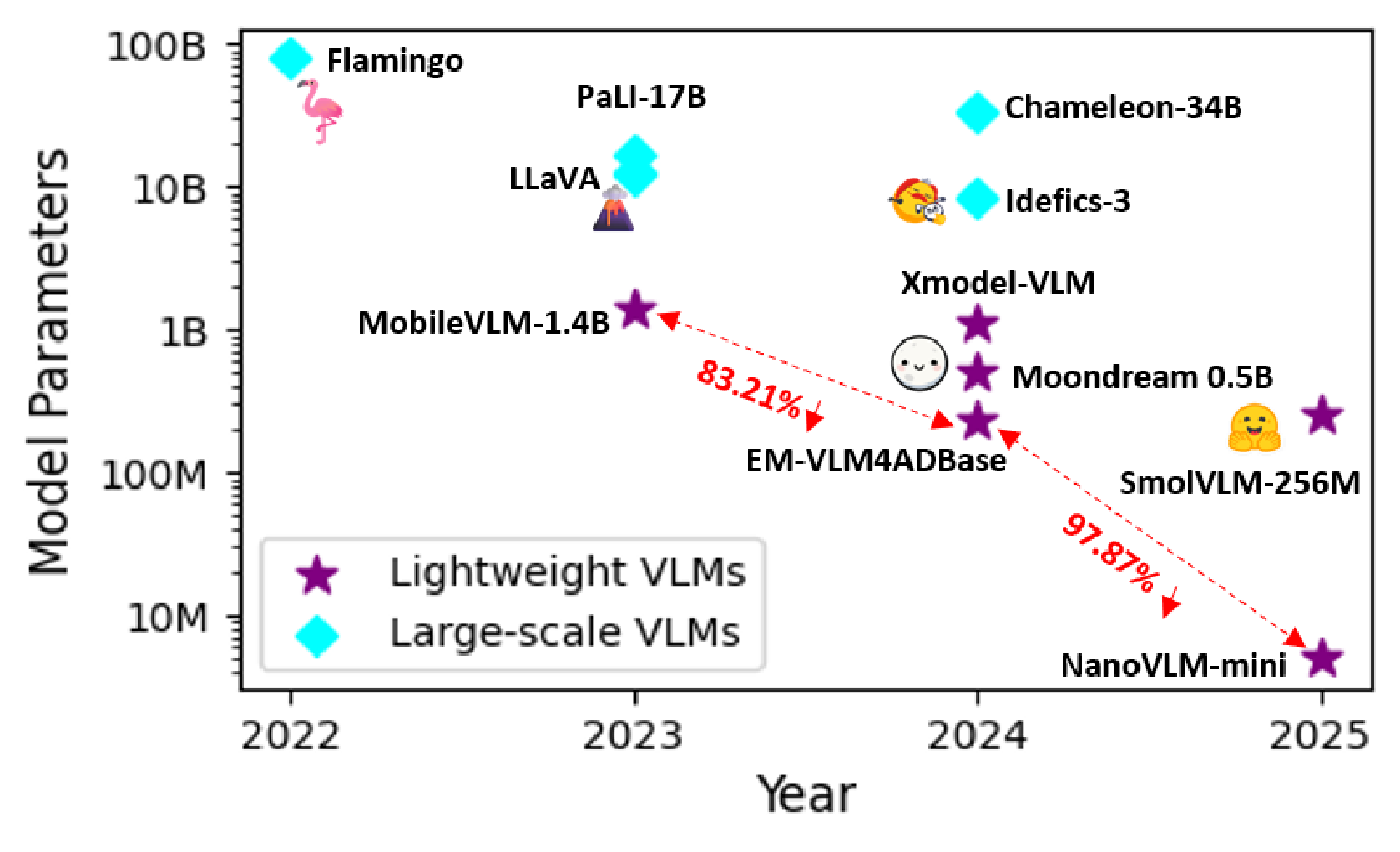}
    \caption{Parameter distribution of some lightweight and large-scale VLMs over time. Recent years have witnessed a growing emphasis on lightweight VLMs.}
    \label{fig:LightweightVLMs}
\end{figure}

The need for efficiency, scalability, and deployment on resource-constrained devices has driven the transition from large-parameter VLMs to lightweight models.
Smaller VLMs optimize architecture and tokenization to maintain strong multimodal performance while reducing memory, latency, and energy costs. Lightweight models usually reduce the number of parameters using techniques or a combination of techniques described above. Figure \ref{fig:LightweightVLMs} compares model parameters between some lightweight and large-scale VLMs.

\subsection{SmolVLM}
SmolVLM-256M and SmolVLM-500M \cite{marafioti2025smolvlm} are the latest additions to the lightweight VLM family. SmolVLM-256M performs well in captioning, document Q\&A, and basic visual reasoning tasks. SmolVLM-500M offers notable performance gains over its sibling while remaining lightweight. The latest models achieve this efficiency by replacing the SigLIP 400M SO vision encoder with the SigLIP base patch-16/512 (93M).

\subsection{MobileVLM}
MobileVLM \cite{Chu2023MobileVLMA} is a high-performance, mobile-scale VLM designed for efficient deployment on resource-constrained devices. It introduces MobileLLaMA (1.4B \& 2.7B parameters), along with an optimized vision encoder and a lightweight downsample projector (LDP) for efficient visual-text alignment. MobileVLM offers strong multimodal performance on various benchmarks while maintaining low latency, with 21.5 tokens/s on Snapdragon 888 CPU and 65.3 tokens/s on Jetson Orin GPU.

\subsection{EM-VLM4AD}
EM-VLM4AD \cite{gopalkrishnan2024EM-VLM4AD} is a lightweight VLM developed for  VQA in autonomous driving. It efficiently processes multi-view traffic scene images using gated pooling attention to extract relevant spatial information. The model features two lightweight language backbones: a fine-tuned T5-Base (235M parameters) and an 8-bit quantized T5-Large (769M parameters), significantly reducing memory (up to 10x) and FLOP requirements compared to prior VLMs.

\subsection{EfficientVLM}
EfficientVLM \cite{WangZ0Z23EfficientVLM} is a fast and compact VLM that reduces parameter count to 93M (44.3\% of the teacher model) while retaining 98.4\% performance and accelerating inference by 2.2×. This is achieved by a distillation then pruning approach, wherein knowledge distillation compresses the model while maintaining accuracy, and modal-adaptive pruning selectively eliminates redundant structures in vision and language encoders according to task importance.

\subsection{MiniVLM} 
MiniVLM \cite{wang2021minivlmsmallerfastervisionlanguage} is a lightweight VLM that reduces parameters to 27\% and FLOPs to 1\% of OSCAR \cite{Li2020OSCAR} while retaining 94–97\% accuracy. This makes MiniVLM optimal for resource-constrained deployments . It achieves this via TEE (Two-stage Efficient feature Extractor) for faster visual processing and a MiniLM-based transformer for vision-language fusion.

\subsection{LightCLIP}
LightCLIP \cite{nie2023lightcliplearningmultilevelinteraction} enhances efficiency with three key techniques: \blackcircled{1} progressive label softening for better instance alignment, \blackcircled{2} bipartite matching for fine-grained image-text alignment, \blackcircled{3} and MLM with image-text fusion to optimize a compact text encoder. It significantly improves zero-shot classification and retrieval performance without increasing inference costs, making it useful for resource-constrained environments.

\subsection{NanoVLMs}
NanoVLMs \cite{agarwalla2025nanovlms} introduce ultra-small VLMs (5M, 16M, \& 25M parameters) optimized for efficiency. NanoVLMs prioritize  better parameter allocation across a compact visual encoder, a lightweight language decoder, and a simple visual-text connector. The novelty lies in scaling down VLMs without sacrificing coherence or fluency.

\subsection{Moondream}
Moondream \cite{moondream2025} 2B (1.9B parameters) supports fp16, int8, and int4 quantization for efficient GPU and CPU inference on mobile devices. Moondream 0.5B (0.5B parameters) uses int8 and int4 quantization and is optimized for mobile and edge devices.  Both models use QAT (refer to section \ref{QAT}) to enhance efficiency.

\subsection{Xmodel-VLM}
Xmodel-VLM \cite{xu2024xmodelvlm} is a 1B-scale lightweight VLM which uses CLIP ViT-L/14@336 for vision, Xmodel-LM-1B for language, and an XDP projector for reduced visual tokens. Following the LLaVA paradigm, it goes through two stages of training for feature alignment and instruction tuning. Despite being small, it delivers competitive performance on multimodal benchmarks while maintaining low inference latency.

Furthermore, Table \ref{tab:frameworks} outlines various frameworks and libraries and highlights key features of each.

\begin{table}[!htbp]
\centering
\caption{Comparison of \textcolor{blue}{frameworks} \& \textcolor{cyan}{libraries} for VLM optimization, fine-tuning, edge execution, and FL integration. \cmark indicates support, \xmark indicates no support, and \halfcirc indicates partial support.}
\label{tab:frameworks}
\scalebox{0.85}{
    \scriptsize
    \setlength{\tabcolsep}{2pt}  
    \begin{tabular}{|C{2cm}|C{3cm}|C{3cm}|C{2cm}|C{3cm}|C{2cm}|}
    \hline
    \textbf{Framework / Library} & \textbf{Optimization Techniques Supported} & \textbf{Fine-Tuning Techniques Supported} & \textbf{Edge Execution} & \textbf{Supported VLM Models} & \textbf{Federated Learning Support} \\
    \hline
    \textcolor{blue}{EdgeVL} \cite{edgevl} & Dual-modality knowledge distillation, quantization-aware training & Fine-tuning via contrastive learning and distillation & \cmark & Adapted CLIP-based VLMs for edge deployment & \xmark \\
    \hline
    \textcolor{cyan}{DeepSpeed} \footnotemark[1] & Zero and sparse training optimizations & Distributed fine-tuning and model parallelism & \cmark & Large VLMs (e.g., LLaVA variants, Flamingo, MiniGPT-4, etc.) & Not directly \\
    \hline
    \textcolor{cyan}{Hugging Face Transformers \footnotemark[2] + Optimum \footnotemark[3]} & Quantization (INT8/FP16), pruning, distillation & Full fine-tuning, LoRA, adapters, prompt tuning & \cmark & CLIP, BLIP, ViLT, VisualBERT, FLAVA, LLaVA variants (e.g., LLaVA, LLaVA 1.5, LLaVA 1.6, LLaVA-CoT, LLaVA-Plus, u-LLaVA, etc.) & \halfcirc \\
    \hline
    \textcolor{cyan}{TensorRT-LLM} \footnotemark[4] & INT8 \& FP16 quantization, layer fusion & Inference-only (no native fine-tuning) & \cmark & Converted models (e.g., CLIP, BLIP, MobileVLM) & \xmark \\
    \hline
    \textcolor{blue}{OpenVINO (with NNCF)} \footnotemark[5] & Post-training quantization, structured pruning, sparsity & Limited transfer learning via model conversion & \cmark & CLIP, custom compact VLMs (e.g., MiniVLM ports) & \xmark \\
    \hline
    \textcolor{blue}{Apache TVM} \footnotemark[6] & Quantization and pruning & Primarily a compiler/optimization tool & \cmark & Any exportable VLM (e.g., CLIP, BLIP, Flamingo variants) & \xmark \\
    \hline
    \textcolor{cyan}{ONNX Runtime} \footnotemark[7] & Graph optimizations, quantization & Emerging support for ONNX training (minimal fine-tuning) & \cmark & VLMs (e.g., CLIP, BLIP, MobileVLM, LLaVA variants) & \xmark \\
    \hline
    \textcolor{blue}{NVIDIA TAO Toolkit} \footnotemark[8] & Quantization, pruning, distillation & Transfer learning with parameter-efficient methods (e.g., adapters) & \cmark & NVIDIA-optimized models (e.g., MobileVLM V2, Moondream2) & \xmark \\
    \hline
    \textcolor{blue}{NVIDIA Triton Inference Server} \footnotemark[9] & Graph optimizations, quantization, runtime optimizations & N/A (serves for inference, not fine-tuning) & \cmark & VLMs (e.g., CLIP, BLIP, MobileVLM, LLaVA variants) & \xmark \\
    \hline
    \textcolor{cyan}{FedML} \footnotemark[10] & Via integration with standard compression libraries & Federated fine-tuning, including parameter-efficient approaches & \cmark & Custom VLMs (e.g., CLIP variants) via federated training & \cmark \\
    \hline
    \textcolor{cyan}{PySyft} \footnotemark[2] & Can be integrated with standard compression techniques & Federated fine-tuning with secure aggregation & \cmark & Custom PyTorch-based VLMs (e.g., CLIP, BLIP, LLaVA variants) & \cmark \\
    \hline
    \end{tabular}
}
\end{table}

\section{Analysis and Insights}
\label{insights}
In this section, we analyze the impact of efficiency-driven techniques on accuracy, latency, and memory consumption. Our evaluation is based on two models: \textit{\blackcircled{1} blip-vqa-base \cite{Li2022BLIP}} (\(\sim 385M\) parameters) and \textit{\blackcircled{2} vilt-b32-finetuned-vqa \cite{Kim2021ViLT}} (\(\sim 87M\) parameters). All results are reported using an NVIDIA RTX A6000 GPU. Notably, the observed trends remain consistent across all models, provided the parameter count falls within the range of these two models. This section provides insights specifically for pre-deployment techniques such as quantization, pruning, and low-rank approximation. We focus on these methods because they are fundamental for optimizing models before deployment, directly impacting model size and performance. Although efficient fine-tuning, runtime optimizations, and privacy-preserving methods are important, they are often model- or task-specific and involve different challenges and considerations that extend beyond the scope of this section. Our analysis provides insights into the accuracy vs. efficiency trade-off, emphasizing the importance of selecting the right technique based on the specific requirements of the downstream application.

\footnotetext[1]{\url{https://www.deepspeed.ai/}}
\footnotetext[2]{\url{https://github.com/OpenMined/PySyft}}
\footnotetext[3]{\url{https://huggingface.co/docs/optimum/}}
\footnotetext[4]{\url{https://developer.nvidia.com/tensorrt}}
\footnotetext[5]{\url{https://docs.openvino.ai/2025/index.html}}
\footnotetext[6]{\url{https://tvm.apache.org/}}
\footnotetext[7]{\url{https://github.com/microsoft/onnxruntime}}
\footnotetext[8]{\url{https://developer.nvidia.com/tao-toolkit}}
\footnotetext[9]{\url{https://github.com/triton-inference-server/server}}
\footnotetext[10]{\url{https://fedml.ai/}}

\begin{figure}[htbp]
    \centering
    \includegraphics[width=\textwidth]{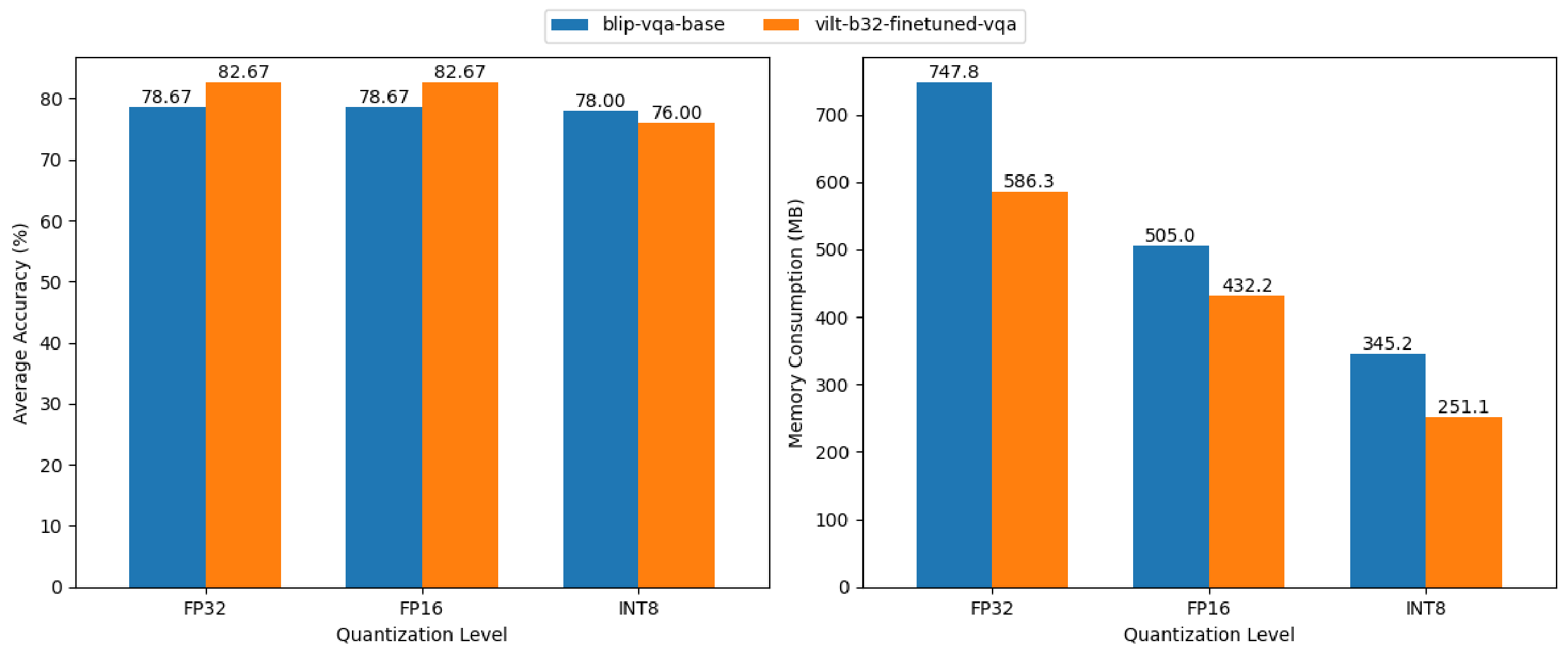}
    \caption{(Left) Average accuracy for FP32, FP16, and INT8 quantization across 50 samples. (Right) Corresponding memory consumption.}
    \label{fig:quantization_memory_accuracy}
\end{figure}

Quantization (figure \ref{fig:quantization_memory_accuracy}) has a notable impact on accuracy, latency, and memory consumption on both blip-vqa-base and vilt-b32-finetned-vqa. For blip-vqa-base, transitioning from FP32 to INT8 results in a 0.85\% accuracy drop for 50 samples 
whereas vilt-b32-finetned-vqa experiences an 8.07\% reduction  
indicating that ViLT is more sensitive to INT8 quantization.
We also increase the number of samples to 125 to observe the effect on accuracy and memory consumption. Increasing the sample size from 50 to 125 causes a 1.36\% accuracy drop in FP32 and a 1.54\% drop in INT8 for blip-vqa-base, while vilt-b32-finetned-vqa sees a 7.1\% decrease in FP32 and a 0.09\% reduction in INT8. However, the accuracy drop for vilt-b32-finetuned-vqa from FP32 to INT8 is steeper (4.6\% on average) compared to blip-vqa-base (0.94\%), suggesting that vilt-b32-finetuned-vqa benefits more from higher precision, but suffers greater degradation from aggressive quantization. In terms of memory consumption, quantization significantly reduces memory requirements across both models. For blip-vqa-base, moving from FP32 to INT8 results in a 53.8\% reduction (747.8MB to 345.2MB) for 50 samples and a 43.1\% reduction (800MB to 455MB) for 125 samples, while vilt-b32-finetuned-vqa achieves a 57.2\% reduction (586.3MB to 251.1MB) for 50 samples and 44.2\% (664.4MB to 370.6MB) for 125 samples, demonstrating that INT8 provides substantial memory savings, with vilt-b32-finetuned-vqa benefiting slightly more. Across all quantization levels, blip-vqa-base consistently consumes more memory than vilt-b32-finetuned-vqa, with a difference of 161.5MB in FP32 (50 samples) and 94.1MB in INT8 (50 samples). The blip-vqa-base maintains a more stable accuracy across sample sizes but consistently requires more memory. The latency per sample remains 0.0877s, 0.0757s, and 0.0701s across FP32, FP16, and INT8 for blip-vqa-base and 0.0330s, 0.0321s, and 0.0311s for vilt-b32-finetuned-vqa, with INT8 being the fastest in both cases. For FP32, we run the models with full 32-bit precision. For FP16, mixed precision is enabled using autocasting to boost speed while maintaining accuracy. For INT8, a copy is dynamically quantized on its linear layers to shrink the model and speed up inference.

\begin{figure}[htbp]
    \centering
    \includegraphics[width=\textwidth]{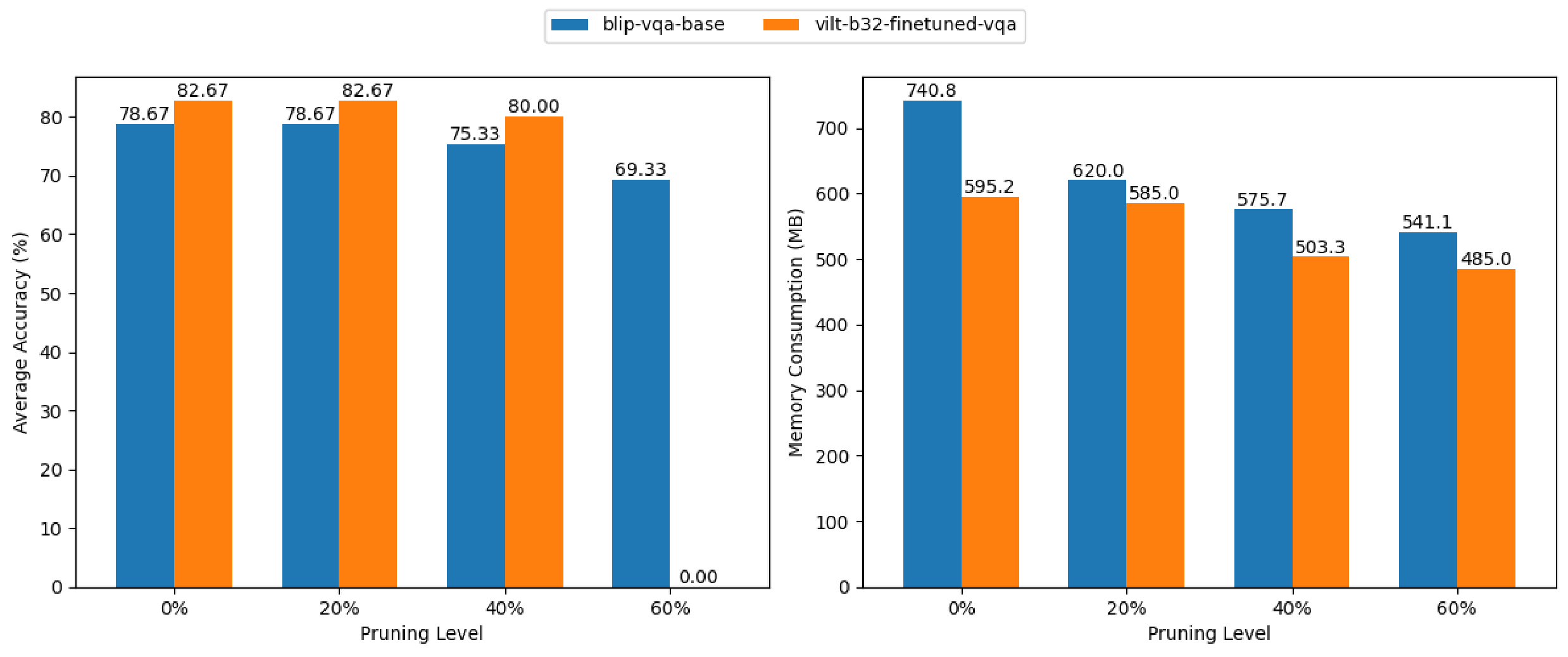}
    \caption{(Left) Average accuracy for 0\%, 20\%, 40\% and 60\% pruning across 50 samples. (Right) Corresponding memory consumption.}
    \label{fig:pruning_memory_accuracy}
\end{figure}

Pruning (figure \ref{fig:pruning_memory_accuracy}) significantly impacts accuracy and memory consumption across both models. For blip-vqa-base, accuracy drops by 0\% at 20\% pruning,  4.24\% at 40\% pruning and 11.9\% at 60\% pruning for 50 samples.
The vilt-b32-finetuned-vqa model follows a similar trend, but at 60\% pruning, accuracy drops to 0\%, indicating extreme sensitivity to high sparsity levels. In terms of memory savings, blip-vqa-base reduces by 16.3\% at 20\% pruning, 22.3\% at 40\% pruning, and 27\% at 60\% pruning (50 samples), while vilt-b32-finetuned-vqa achieves reductions of 1.7\%, 15.4\%, and 18.5\%, respectively. For 125 samples, blip-vqa-base saves 12.3\%, 20.7\%, and 24.9\%, whereas vilt-b32-finetuned-vqa reduces by 4.2\%, 13.4\%, and 21\%. The results show that vilt-b32-finetuned-vqa maintains better accuracy at lower pruning levels but is highly sensitive to aggressive pruning. The latency per sample for blip-vqa-base at 20\%, 40\%, and 60\% pruning is 0.0569s, 0.0561s, and 0.0559s, respectively, while for vilt-b32-finetuned-vqa, it is 0.0148s, 0.0141s, and 0.0139s at the same pruning levels. We observe that pruning results in faster inference per sample compared to quantization. We implement pruning by scanning all model modules to find linear layers and then apply global unstructured L1 pruning with a set sparsity ratio. After pruning, the extra reparameterization is removed, making the pruned model ready for inference.

\begin{figure}[htbp]
    \centering
    \includegraphics[width=\textwidth]{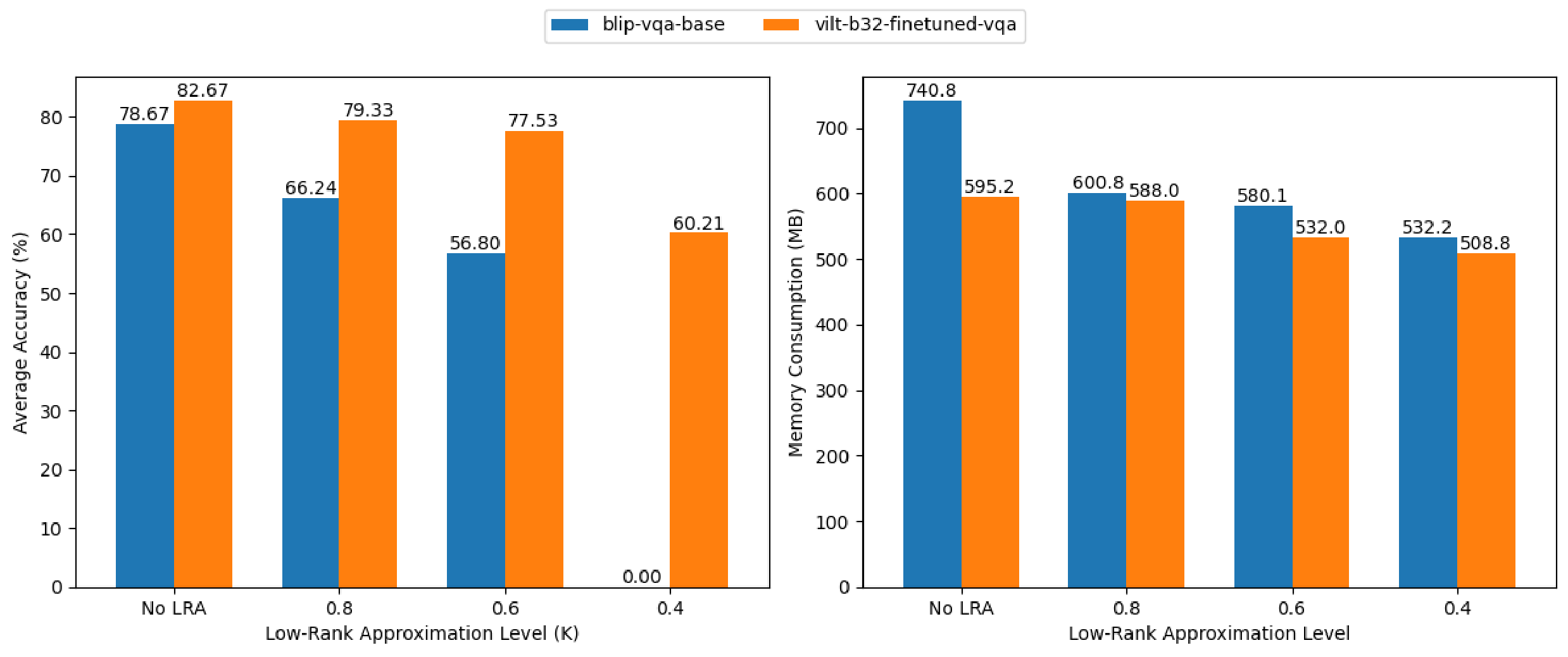}
    \caption{(Left) Average accuracy for no low-rank approximation, k = 0.8, k = 0.6, and k = 0.4 across 50 samples. (Right) Corresponding memory consumption.}
    \label{fig:LRA_memory_accuracy}
\end{figure}

Low-rank approximation (figure \ref{fig:LRA_memory_accuracy}) significantly affects accuracy and memory consumption across both models. Decreasing the rank from k = 0.8 to k = 0.4 increases compression, leading to lower memory consumption and latency. However, this also results in a drop in accuracy. For blip-vqa-base, accuracy decreases by 15.8\% at k = 0.8, 27.8\% at k = 0.6, and drops to 0\% at k = 0.4 for 50 samples. The vilt-b32-finetuned-vqa exhibits a more gradual decline, with accuracy dropping by 4\% at k = 0.8, 6.2\% at k = 0.6, and 27.1\% at k = 0.4, indicating that vilt-b32-finetuned-vqa is more resilient to LRA compared to blip-vqa-base. In terms of memory savings, blip-vqa-base reduces memory consumption by 140 MB at k = 0.8, 160.7 MB at k = 0.6, and 208.6 MB at k = 0.4 (50 samples), while vilt-b32-finetuned-vqa achieves reductions of 7.2 MB, 63.2 MB, and 86.4 MB, respectively. For 125 samples, blip-vqa-base saves 65.4 MB, 100.2 MB, and 199.3 MB, whereas vilt-b32-finetuned-vqa reduces by 64 MB, 83.8 MB, and 150 MB. This indicates that while low-rank approximation provides substantial memory savings, blip-vqa-base is highly sensitive to aggressive rank reduction, whereas vilt-b32-finetuned-vqa maintains better accuracy at even lower ranks. The latency per sample for blip-vqa-base at k = 0.8, k = 0.6, and k = 0.4 is 0.0442s, 0.0435s, and 0.0429s, respectively, while for vilt-b32-finetuned-vqa, it is 0.0148s, 0.0136s, and 0.0136s at the same rank reduction levels. We observe that low-rank approximation leads to lower latency per sample compared to quantization and pruning. We implement low-rank approximation by iterating over each linear layer, computing a reduced rank from a given ratio, and performing SVD on the weight matrix. Only the top singular values and vectors are kept to reconstruct an approximated weight, which replaces the original.

\section{Applications}
\label{applications}
VLMs have demonstrated exceptional performance in tasks such as image captioning, VQA, and document understanding. The development of compact models has extended their applicability to real-time applications, allowing for deployment on resource-constrained devices. These models are utilized across various domains, including robotics, medical applications, autonomous driving, augmented and virtual reality (AR/VR), and smart surveillance. Table \ref{tab:application_datasets} provides datasets to train VLMs for various downstream applications.

\subsection{Autonomous Driving}
VLMs play a crucial role in autonomous driving by enabling robust scene understanding, decision-making, and natural language-based interactions. DriveVLM \cite{Tian2024DriveVLM} enhances autonomous driving by leveraging VLMs for advanced scene understanding and planning. It leverages the chain-of-thought (CoT) process with modules  for scene description, analysis, and hierarchical planning. To address spatial reasoning and computing limits, DriveVLM-Dual combines VLM-based reasoning with traditional 3D perception and  planning, resulting in improved real-time decision-making. DriveVLM demonstrates its effectiveness through successful deployment in real-world scenarios. To address the lack of natural language integration for interpretable decision-making in autonomous driving, \cite{Park2024VLAAD} introduce an instruction tuning dataset and train a multimodal model called VLAAD for enhanced scene understanding and reasoning. V2X-VLM \cite{You2024V2XVLMEV} performs end-to-end vehicle-infrastructure cooperative autonomous driving for enhanced situational awareness and optimized trajectory planning.

\subsection{Medical Applications}
VLMs are used extensively in medical applications for question-answering and medical report generation. They also help interpret medical images by combining visual and textual information for better diagnosis. MedViLL \cite{Moon2022MedViLL} performs vision-language pre-training and multi-modal representation learning for medical applications. It improves diagnostic classification, medical image report retrieval, visual question answering, and radiology report generation with a BERT-based architecture \cite{devlin-etal-2019-bert} and a novel attention masking method. PubMedCLIP \cite{eslami-etal-2023-pubmedclip} fine-tunes CLIP for medical visual question answering (MedVQA) using medical image-text pairs from PubMed. Unlike previous  approaches, it is not limited to certain body regions and  supports several imaging modalities (X-ray, CT, MRI). \textit{SERPENT-VLM} \cite{Kapadnis2024SERPENTVLMS} reduces hallucinations and improves radiology report generation by refining image-text alignment using a self-supervised loss. It outperforms LlaVA-Med and BiomedGPT on IU X-ray and ROCO datasets while maintaining robustness against noisy images. Med-MoE \cite{jiang2024MedMoE} enables lightweight medical VLMs using Mixture-of-Experts, optimizing efficiency for Med-VQA and classification tasks while reducing compute needs.

\begin{table}[!htbp]
\centering
\caption{Datasets with image-text pairs for training VLMs in various applications. Refer to \cite{Zhou2023autonomousdriving, Hartsock_2024medical} for more datasets on autonomous driving and medical applications, respectively. }
\label{tab:application_datasets}
\scalebox{1}{
    \scriptsize
    \setlength{\tabcolsep}{2pt}  
    \begin{tabular}{|C{2cm}|C{2cm}|C{7cm}|C{2.5cm}|}
    \hline
    \textbf{Dataset} & \textbf{Year} & \textbf{Key Features} & \textbf{Link} \\
    \hline
    \multicolumn{4}{|c|}{\textcolor{blue}{\large \textit{AUTONOMOUS DRIVING DATASETS}}} \\
    \hline
    BDD-X \cite{kim2018BDD-X} & 2018 & Dataset contains over 26K activities in over 8.4M frames. Train and test set contain 5,597 and 656 videos respectively. Used for action explanation. & \href{https://github.com/JinkyuKimUCB/BDD-X-dataset}{\includegraphics[height=10pt]{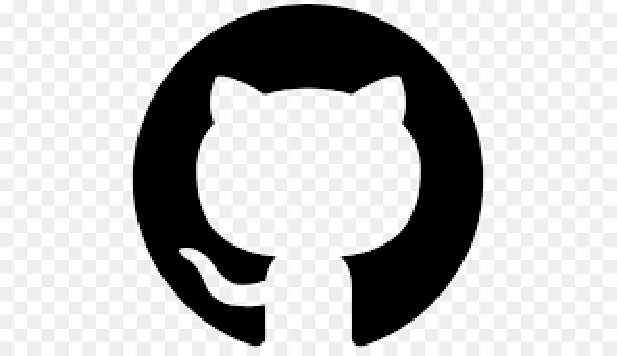}} \\
    \hline
    TOUCHDOWN \cite{Chen2019Touchdown} & 2019 & Used for natural language navigation and spatial reasoning in urban environments. 6,526 training, 1,391 development, 1,409 test for navigation; 17,880 training, 3,836 development, 3,859 test for SDR. & \href{https://github.com/lil-lab/touchdown}{\includegraphics[height=10pt]{Tex/images/github.eps}} \\
    \hline
    HAD \cite{kim2019CVPRHAD} & 2019 & Contains 30 hours of driving data with natural language advice. Primarily used for Human-to-Vehicle Advice. & \href{https://usa.honda-ri.com/had}{\includegraphics[height=10pt]{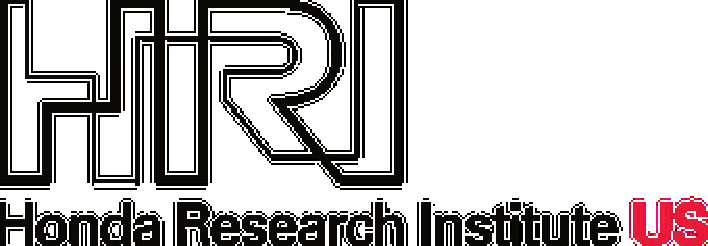}} \\
    \hline
    Talk2Car \cite{deruyttere2019talk2car} & 2020 & Provides free-form natural language commands for autonomous driving. Built on \textit{nuScenes}, with multimodal sensor data (RGB, LIDAR, RADAR, GPS, semantic maps, 3D bounding boxes). It has 8349 train samples and 2447 test samples. & \href{https://github.com/talk2car/Talk2Car}{\includegraphics[height=10pt]{Tex/images/github.eps}} \\
    \hline
    CARLA-NAV \cite{jain2022CARLA-NAV} & 2022 & It has video sequences from 8 maps and 14 weather conditions. It contains dynamic scenes and trajectory annotations for real-time navigation. & \href{https://github.com/kanji95/Carla-Nav-Tool}{\includegraphics[height=10pt]{Tex/images/github.eps}} \\
    \hline
    Talk2BEV \cite{talk2bev} & 2023 & 1000 human-annotated BEV scenarios with 20,000+ questions and ground-truth answers from the \textit{nuScenes} dataset. Used for open-loop decision-making. & \href{https://github.com/llmbev/talk2bev}{\includegraphics[height=10pt]{Tex/images/github.eps}} \\
    \hline
    DRAMA \cite{malla2023drama} & 2023 & 17,785 urban traffic clips from Tokyo with synchronized video, CAN signals, and IMU data, featuring annotations like Q/A, bounding boxes, and driver suggestions. & \href{https://usa.honda-ri.com/drama}{\includegraphics[height=10pt]{Tex/images/HRI.eps}} \\
    \hline
    \multicolumn{4}{|c|}{\textcolor{blue}{\large \textit{MEDICAL DATASETS}}} \\
    \hline
    ROCO \cite{ROCO} & 2018 & A large-scale medical imaging dataset featuring radiology and non-radiology images with captions, keywords, UMLS Semantic Types, and Concept Unique Identifiers (CUIs), useful for image captioning, classification, and retrieval. & \href{https://github.com/razorx89/roco-dataset}{\includegraphics[height=10pt]{Tex/images/github.eps}} \\
    \hline
    MIMIC-CXR \cite{MIMIC-CXR}& 2019 & 377,110 chest X-ray images in DICOM format with corresponding free-text radiology reports. It includes patient and study identifiers, structured metadata, and de-identified annotations for research in medical imaging. & \href{https://physionet.org/content/mimic-cxr/2.1.0/}{\includegraphics[height=10pt]{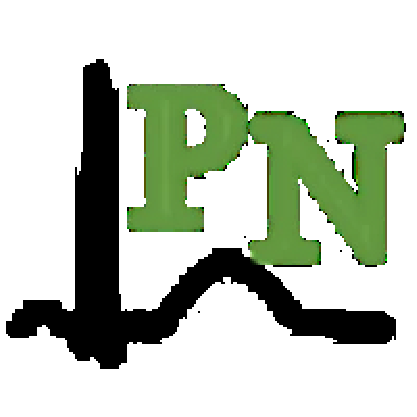}} \\
    \hline
    MIMIC-CXR-JPG \cite{MIMIC-CXR-JPG, MIMIC-CXR-JPG-ArXiv} & 2019 & MIMIC-CXR-JPG uses JPG format with structured labels, while MIMIC-CXR provides DICOM images with free-text reports. & \href{https://physionet.org/content/mimic-cxr-jpg/2.1.0/}{\includegraphics[height=10pt]{Tex/images/physionet.eps}} \\
    \hline
    MedICaT \cite{subramanian-2020-medicat} & 2020 & Contains 217,060 medical figures with captions, subfigure annotations, and inline references from PubMed Central and S2ORC. & \href{https://github.com/allenai/medicat}{\includegraphics[height=10pt]{Tex/images/github.eps}} \\
    \hline
    PMC-OA \cite{Lin2023PMC-CLIP} & 2023 & Biomedical dataset with 1.65M image-caption pairs. Commonly used for retrieval, classification, and VQA tasks. & \href{https://huggingface.co/datasets/axiong/pmc_oa}{\includegraphics[height=10pt]{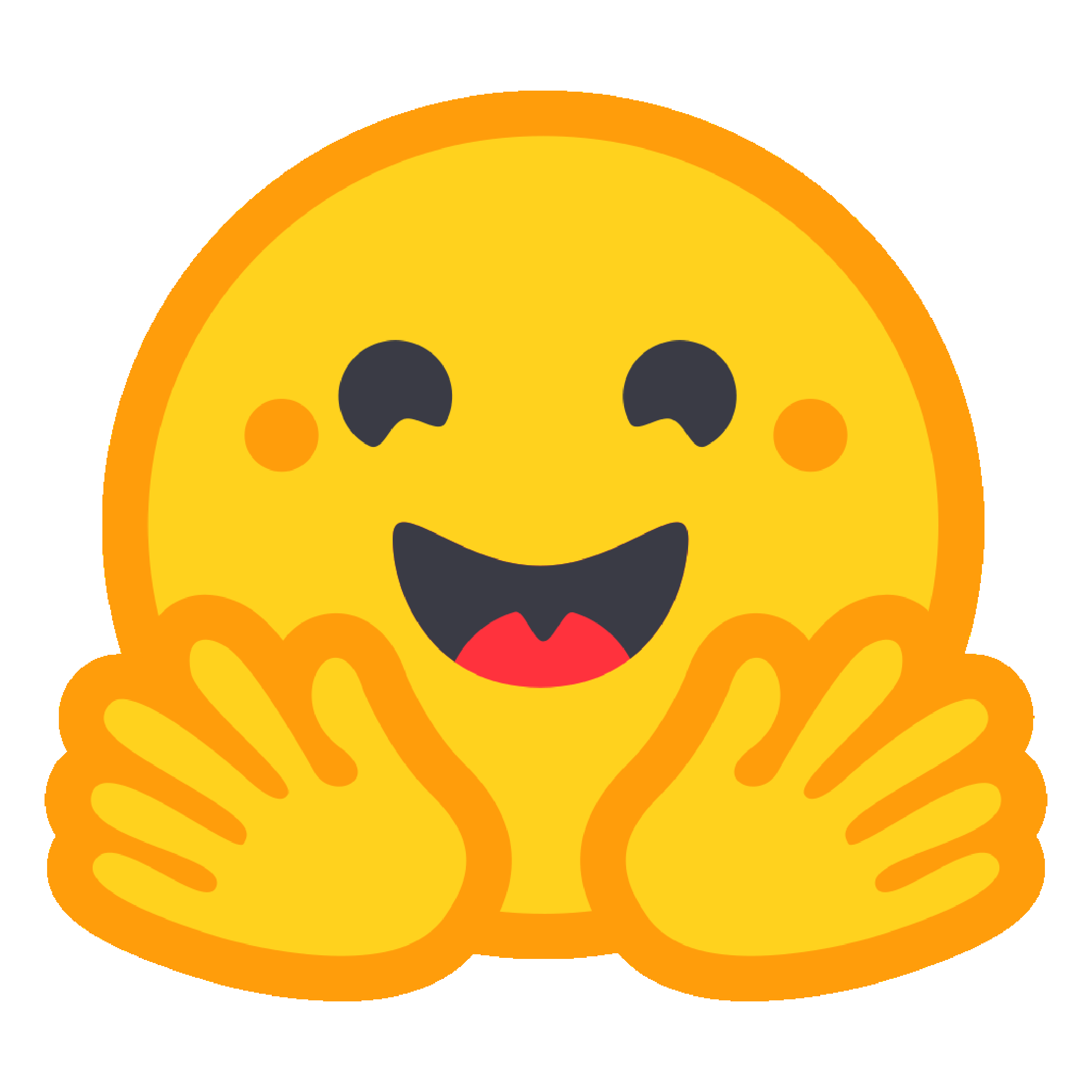}} \\
    \hline
    \multicolumn{4}{|c|}{\textcolor{blue}{\large \textit{ROBOT NAVIGATION \& MANIPULATION DATASETS}}} \\
    \hline
    CC3M \cite{sharma-etal-2018-conceptual} & 2018 & Contains 3 million image-text pairs collected from the web. Helps robots understand visual scenes through language grounding. & \href{https://ai.google.com/research/ConceptualCaptions/}{Webpage} \\
    \hline
    MQA \cite{deng2020mqa} & 2021 & Simulation-based dataset for manipulation question answering (MQA). It includes scene variations, Q\&A pairs, and real-world transferability. & \href{https://github.com/dengyh16code/MQA_dataset}{\includegraphics[height=10pt]{Tex/images/github.eps}} \\
    \hline
    ManipVQA \cite{huang2024manipvqainjectingroboticaffordance} & 2024 & It introduces a VQA format for robotic affordance understanding by integrating tool detection, affordance recognition, and physical concept grounding. & \href{https://github.com/SiyuanHuang95/ManipVQA}{\includegraphics[height=10pt]{Tex/images/github.eps}} \\
    \hline
    PHYSOBJECTS \cite{Gao2024pysobjects} & 2024 & Object-centric dataset with 39.6K crowd-sourced and 417K automated physical concept annotations for household objects. Used for robotic manipulation \& planning. & \href{https://iliad.stanford.edu/pg-vlm/}{Webpage} \\
    \hline
    \multicolumn{4}{|c|}{\textcolor{blue}{\large \textit{SMART SURVEILLANCE DATASETS}}} \\
    \hline
    COYO-700M \cite{kakaobrain2022coyo-700m} & 2022 &  747M image-text pairs useful for scene understanding and anomaly detection & \href{https://github.com/kakaobrain/coyo-dataset}{\includegraphics[height=10pt]{Tex/images/github.eps}} \\
    \hline
    FaceCaption-15M \cite{dai202415facecaption} & 2024 & Useful for smart surveillance and security as it provides detailed facial image-text pairs, enabling advanced facial recognition, identity verification and suspect tracking. & \href{https://huggingface.co/datasets/OpenFace-CQUPT/FaceCaption-15M}{\includegraphics[height=10pt]{Tex/images/huggingface.eps}} \\
    \hline
    \end{tabular}
}
\end{table}

\subsection{Robotic Navigation and Manipulation}
Recently, considerable efforts have been directed toward integrating VLMs for robotic navigation and manipulation, which are crucial for search and rescue (SaR) operations. Effective robotic navigation and manipulation in challenging environments require a precise understanding of physical properties and geometric relationships. Geometric awareness includes assessing whether an object can fit inside another or identifying appropriate cover, enabling robots to make informed decisions and interact safely with their surroundings. Physical grounding is essential for search and rescue operations and hazard detection. \cite{Gao2024PhysObjects} introduces \textit{PHYSOBJECTS}, a dataset incorporating physical concepts like mass, fragility, deformability, and density. They fine-tune a VLM on this dataset and propose a robotic manipulation framework where the LLM queries the VLM, receives responses, and formulates an execution plan. VLM-GroNav \cite{elnoor2024VLMGRONAV} integrates a VLM-based reasoning module with physical grounding to evaluate terrain properties like slipperiness and deformability. The framework is tested on two robots: Ghost Vision 60 \footnote[11]{\url{https://www.ghostrobotics.io/vision-60}} and Clearpath Husky \footnote[12]{\url{https://clearpathrobotics.com/husky-a300-unmanned-ground-vehicle-robot/}}. \cite{ahmad2024addressingfailuresroboticsusing} integrate VLMs with Behavior Trees (BTs) to detect, identify, and recover from failures in robotic tasks. The VLM analyzes visual input to recognize failures, suggest missing conditions and update the BT to handle similar failures in the future. VLM-Social-Nav \cite{sONG2025vlmsOCIALNAV} leverages VLMs to interpret social context and compute a navigation cost, enabling robots to make socially aware decisions in human environments.

\subsection{Smart Surveillance}
VLMs have transformed smart surveillance by enabling advanced scene understanding and contextual reasoning. \cite{Li2024SmartCity} introduce a FL framework for domain-specific VLMs, enabling collaborative fine-tuning of large multimodal models for smart city safety and urban operation management. \cite{desilva2025largelanguagemodelsvideo} introduces a novel VLM-based pipeline for efficient video analysis. They reduce storage space by replacing video data with detailed textual summaries from CCTV feeds. Video-ChatGPT \cite{maaz-etal-2024-video} provides detailed video understanding and conversational AI capabilities, which can be integrated for smart surveillance by fine-tuning it on security footage for real-time anomaly detection and event summarization.

\subsection{Augmented Reality (AR)}
VLMs have emerged as an effective tool for enhancing augmented reality experiences. ViDDAR \cite{xiu2025viddar} detects obstruction and information manipulation attacks for enhancing safety and usability in AR. It uses VLMs to analyze virtual content interactions and assess their impact on real-world scene understanding. ViDDAR employs a user-edge-cloud architecture to balance detection accuracy and latency. \cite{duan2025DiverseAR} introduce DiverseAR, the first dataset designed to evaluate the capability of VLMs to identify and describe virtual content at varying levels of AR scene complexity. TAGGAR \cite{Stover2024TAGGAR} is a general-purpose AR task guidance system that uses VLMs to generate automated visual instructions from natural language and images. It eliminates the need for expert-placed AR visuals or CAD models by allowing operators to specify tasks with text and images.

\section{Open Challenges and Future Directions}
\label{challenges}
Despite the proliferation of Vision-Language Models (VLMs) in recent months, their deployment on resource-constrained devices for real-time applications remains a substantial challenge. A primary limitation arises from the computational complexity and large feature representations required to effectively align visual and textual modalities. These high-dimensional embeddings demand significant memory and processing resources, which are often unavailable on edge devices. We outline the key challenges in adapting VLMs for such environments.

\subsubsection{Accuracy Vs Efficiency Tradeoff}

While foundational VLMs can provide great accuracy for various tasks, they are inefficient in terms of memory and computational requirements. Techniques such as quantization and pruning were developed to fit the VLMs on the edge \cite{tinychat-vlm_Acc_Eff}. Matryoshka Quantization (MatQuant) \cite{nair2025matryoshkaquantization} improves efficiency by enabling a single model to operate at multiple precision levels, achieving up to 10\% higher accuracy for int2 models than standard Quantization-Aware Training. 

Benchmarking results have shown that large-scale Vision-Language Models (VLMs) achieve high accuracy across a wide range of tasks and domains \cite{yue2024mmmu_vlm_Acc_Eff}. Although researchers are also developing small VLM models (sub-billion and <3B parmaters) for multimodal multidomain tasks \cite{SmolVLM_Acc_Eff} \cite{zhou2024tinyllava_Acc_Eff} \cite{InternVL2_Acc_Eff}, the architecture and training data plays an important role in using benchmarking dataset to test the efficacy of the model. However, this raises a critical question: \textit{Must edge-deployed VLMs achieve state-of-the-art performance across all tasks?} In practice, VLMs on edge devices are often designed to perform narrow, task-specific functions with high efficiency and accuracy. The real challenge lies in developing specialized VLMs tailored for specific edge applications, which often require compact architectures and context-aware capabilities.

While techniques such as knowledge distillation, low-rank adaptation (LoRA), and adapter-based tuning offer promising solutions for compressing and specializing large models, they remain limited in their ability to capture the full complexity and adaptability needed for task-specific reasoning on resource-constrained hardware \cite{wang2025visualprm_cohesive}. Bridging this gap calls for continued advances in task-aware model compression, self-supervised learning at the edge, and incremental adaptation mechanisms capable of responding to environmental feedback in real time.

\subsubsection{Agentic AI and Distributed VLMs @Edge}

As artificial intelligence systems become increasingly embedded in dynamic, real-world environments, the need for agents capable of perceiving, reasoning, and acting autonomously has intensified \cite{Acharya2025AgenticAI}. However, deploying these capabilities in the real world, especially in settings like robotics, autonomous vehicles, and smart infrastructure, requires moving beyond centralized, cloud-based inference. This leads to the emerging paradigm of distributed VLMs at the edge, where multiple lightweight or specialized models collaborate across heterogeneous devices and contexts \cite{tallam2025AgenticAI}. Multi-agent systems may use shared or federated VLMs to jointly interpret scenes, coordinate tasks, and share knowledge representations \cite{yang2025AgenticAI}. This gives rise to a requirement for a modular architecture where agents may consist of distributed components (e.g., local perception + cloud-based planning), or fully edge-deployed systems with swappable VLM modules based on task requirements. Distributed agents must coordinate semantic knowledge across bandwidth-limited links while maintaining consistent interpretations of multimodal inputs \cite{zhu2025AgenticAI}. Agents must adapt their VLMs to new tasks, domains, and environmental changes without catastrophic forgetting or overfitting. This also calls for Vision encoders and task-specific decoders to share generic features between agents and use task-specific decoders at the receiver to extract meaningful information for different tasks. One such example includes collecting RGB, LiDAR, and IMU information from robots to (a) extract the type of surface using LiDAR and IMU, (b) use RGB and LiDAR for object detection and classification, and (c) use RGB and IMU to detect deformities on the surface.

\subsubsection{On-the-fly Fine-Tuning of VLMs}

On-the-fly fine-tuning methodologies aim to dynamically adapt VLMs to new data or environments without full retraining \cite{malladi2024OntheFly}. These methods are critical for applications in robotics, autonomous systems, wearable devices, and mobile platforms, where VLMs must evolve with changing tasks and environments\cite{tang2024Onthefly}. Some of the existing techniques include meta-learning and test-time adaptation. Meta-learning involves training models to quickly adapt to novel tasks using limited examples. In the VLM domain, meta-learning is instrumental in facilitating few-shot learning, cross-domain generalization, and personalized user adaptation, particularly in low-resource scenarios \cite{najdenkoska2023_OnTheFly} \cite{math12020286_OnTheFly} \cite{hu2023metalearning_OnTheFly}. Test-time adaptation enables models to self-adjust during inference using only unlabeled test samples \cite{karmanov2024_OnTheFly} \cite{farina2024_OnThFly} \cite{NEURIPS2024_TTA}. Unlike meta-learning, TTA does not require episodic task design or access to training data. It is particularly suitable for dynamic, real-time environments where domain shifts or evolving user contexts occur frequently. However, these techniques are still immature to be adapted for edge scenarios. Some of the immediate challenges include: (a) designing lightweight and scalable adaptation frameworks for edge deployment with the above-mentioned techniques; (b) developing unified multimodal task abstractions for few-shot and episodic learning.

\subsubsection{Multimodal VLM @Edge}

While Vision-Language Models (VLMs) are often referred to as multimodal large language models, their modality coverage is typically limited to vision and language, with some models extending to audio or basic spatial context. However, there remains significant untapped potential in broadening the multimodal spectrum to include diverse sensor data, which could substantially enhance model accuracy, contextual understanding, and real-time decision-making across various application domains.

Recent advancements in human activity recognition (HAR) and affective computing suggest that integrating additional sensory modalities can complement visual and linguistic inputs, providing rich, context-aware representations. For instance: Inertial Measurement Units (IMUs) and earable sensors offer continuous, low-latency streams of motion and physiological data that are especially valuable in recognizing fine-grained human activities. Their integration into VLM frameworks could help models distinguish between visually similar actions that differ in physical dynamics (e.g., jogging vs. limping). Millimeter-Wave Radar (mmWave) sensors have been shown to accurately detect human posture and micro-gestures without relying on cameras. In the mmCLIP framework \cite{Cao2024mmCLIP}, radar signals are aligned with vision-language embeddings to improve privacy-preserving human activity recognition, offering robustness in low-light or occluded environments where cameras fail. Electroencephalography (EEG) signals provide a direct interface to brain activity, enabling models to infer cognitive and emotional states in real-time. Recent work \cite{li2024realmindEEG} has integrated EEG with VLM for visual captioning. Audio signals remain a powerful yet underutilized modality. Sonic-VisionLM \cite{Xie2024Sonic_VisionLM} showcases how integrating audio with visual data enhances temporal grounding in video understanding. For instance, matching sounds of footsteps, speech, or environmental cues with corresponding visual scenes allows for more accurate event recognition and scene parsing.

These examples demonstrate the value of moving beyond the traditional image-text paradigm toward truly multimodal VLM architectures. Integrating such diverse sensor modalities can enrich semantic representation, improve generalization, and enable robust performance in complex, real-world scenarios—particularly in wearable computing, robotics, autonomous vehicles, and assistive technologies.

\subsubsection{Reasoning for Cohesive Environment}
The integration of Human-in-the-Loop strategies with Vision-Language Models presents promising opportunities for enhancing the performance, adaptability, and trustworthiness of AI systems in real-world applications to build a cohesive environment. However, this integration introduces several research challenges that must be addressed to ensure effective and reliable deployment. 

\textbf{Model Interpretability and Transparency:} VLMs typically function as black-box systems, making it difficult to understand the reasoning behind their predictions. This lack of interpretability reduces user trust and limits the value of human feedback. Explainable AI techniques such as visual attention maps \cite{selvaraju2017gradcam}, natural language rationales \cite{camburu2018esnli}, and counterfactual reasoning \cite{goyal2019counterfactual} are being explored to bridge this gap. Nonetheless, robust and scalable interpretability methods tailored to multimodal inputs remain an open research problem.
    
\textbf{Confidence Calibration and Uncertainty Estimation:} Human-in-the-loop systems depend on the model’s ability to recognize its own uncertainty and defer to human input when appropriate. Poorly calibrated confidence scores can lead to either over-reliance on the model or unnecessary human intervention. Methods such as temperature scaling \cite{guo2017calibration}, Bayesian neural networks \cite{kendall2017bayesian}, and Monte Carlo Dropout \cite{gal2016dropout} have been proposed to improve uncertainty estimation in deep learning, but their application in VLMs, especially under dynamic multimodal conditions, is still evolving.

\textbf{Bias, Fairness, and Trust in Human Feedback:} Incorporating human feedback does not guarantee fairness; in fact, it may introduce or reinforce biases. Research into adversarial human feedback, trust calibration mechanisms \cite{Okamura2020cohesive}, and feedback quality filtering aims to mitigate these risks. Ensuring equitable and ethical behavior in Human-in-the-loop-VLM systems \cite{Duan2024cohesive} \cite{mcgrath2024cohesive} requires robust governance frameworks and transparent user interaction logging.  

\subsubsection{Security and privacy} 
When VLMs are deployed on end devices, they become vulnerable to adversarial attacks such as Jacobian-based Saliency Map Attack (JSMA) \cite{Wiyatno2018MaximalJS}, which can manipulate model predictions through carefully crafted perturbations. To mitigate these threats, adversarial defense techniques, such as adversarial training \cite{Zhang2024ADVpt} are employed to enhance model resilience against such attacks. In a federated setting, shared gradients are susceptible to inversion attacks \cite{Zhu2019GradientInversion}. These attacks reconstruct private data and compromise privacy. To counteract this, techniques such as differential privacy and gradient pruning \cite{Zhu2019GradientInversion} are used. Differentially Private Federated Prompt Learning (DP-FPL) \cite{tran2025DP-FPL} balances personalization, generalization, and privacy in federated learning for multimodal LLMs. It uses low-rank factorization with a residual term to preserve expressiveness while applying both global and local differential privacy to mitigate privacy risks without significantly degrading performance. However, an honest-but-curious server may still analyze shared updates to infer patterns. Future research should address these challenges to improve the robustness of VLMs in distributed environments. Secure encryption is crucial for cooperative perception tasks in dynamic autonomous systems to prevent adversarial interceptions. AES-256 enables privacy-preserving data transmission. However, with the increasing use of VLMs in distributed applications such as robotics, lightweight and robust encryption algorithms are required to ensure secure information transfer while minimizing computational overhead. 

\section{Conclusion}
\label{conclusion}
Vision-Language Models (VLMs) have rapidly advanced in recent years, demonstrating impressive capabilities in multimodal understanding, visual question answering, image captioning, and grounded reasoning. In this survey, we present a comprehensive and structured review of efficient vision-language models (VLMs). We begin by motivating the need for optimizing VLMs for edge-centric applications and subsequently examine key techniques including pre-deployment strategies, fine-tuning methodologies, and runtime optimizations. The pre-deployment section encompasses quantization, low-rank approximation, pruning, knowledge distillation, and mixture of experts. For efficient fine-tuning, we provide a detailed overview of parameter-efficient and memory-efficient approaches. The runtime optimization section discusses token reduction techniques and test-time adaptation mechanisms. We also detail privacy-preserving distributed VLM approaches, such as federated learning and split learning. Finally, we highlight various lightweight VLMs, discuss their applications, and offer insights into the accuracy-efficiency trade-off of VLMs.

In conclusion, we outline several open challenges in the development and deployment of Vision-Language Models (VLMs) and highlight key questions that remain unanswered within the research community. A fundamental consideration is whether edge-deployed VLMs should be designed for multitasking or optimized for task-specific performance. Depending on the application domain and system constraints, there is a compelling argument for tailoring VLMs to maximize efficiency and accuracy for narrowly defined tasks. Such task-oriented VLMs would benefit from the ability to adapt on-the-fly to dynamic environments, enabling real-time contextual learning and robustness to unforeseen inputs. However, the practical implementation of on-the-fly adaptation techniques remains largely unproven in real-world settings, with few reproducible, evidence-based studies demonstrating their effectiveness at the edge.

Moreover, as VLMs are increasingly adopted in safety-critical and human-centered systems, there are growing concerns regarding their robustness, fairness, and trustworthiness. These models can generate plausible yet incorrect outputs, propagate harmful social biases, or become vulnerable to adversarial inputs, undermining their reliability. Addressing these issues necessitates improved model interpretability, calibrated confidence estimation, and the integration of human-in-the-loop mechanisms to ensure meaningful oversight and corrective intervention. As the field continues to evolve, answering these foundational questions and overcoming the aforementioned limitations will be essential to realizing reliable, adaptive, and ethically aligned multimodal AI systems for deployment in real-world edge and distributed environments.

\section{Conflict of Interest}

The authors have declared no conflicts of interest for this article.


\end{document}